\documentclass{include/latex-class-ecai/ecai}

\usepackage{times}
\usepackage{latexsym}

\usepackage[utf8]{inputenc}
\usepackage{amsmath}
\usepackage{amssymb}
\usepackage{microtype}
\usepackage{bm}
\usepackage{url}
\usepackage{multirow}
\usepackage{listings}
\usepackage{upgreek}
\usepackage{hyperref}
\usepackage{enumerate}
\newtheorem{proposition}{Proposition}
\newtheorem{definition}{Definition}
\newtheorem{corollary}{Corollary}

\newtheorem{lemma}{Lemma}
\newenvironment{proofof}[1]{\noindent {\bf Proof of #1.}}{\qed}

\newcommand{\eqdef}{\ensuremath{\mathbin{\raisebox{-1pt}[-3pt][0pt]{$\stackrel{\mathit{def}}{=}$}}}}
\newcommand{\tuple}[1]{\langle #1 \rangle}

\newcommand{\PV}{\ensuremath{\mathcal{A}}}
\newcommand{\myatom}{\ensuremath{a}}

\newcommand{\HT}{\ensuremath{\mathrm{HT}}}

\newcommand{\LTL}{\ensuremath{\mathrm{LTL}}}

\newcommand{\TEL}{\ensuremath{\mathrm{TEL}}}
\newcommand{\TELf}{\ensuremath{{\TEL}_{\!f}}}

\newcommand{\LDL}{\ensuremath{\mathrm{LDL}}}
\newcommand{\LDLf}{\ensuremath{\mathrm{LDL}_{\!f}}}

\newcommand{\DHT}{\ensuremath{\mathrm{DHT}}}
\newcommand{\DHTf}{\ensuremath{\mathrm{DHT}_{\!f}}}

\newcommand{\DEL}{\ensuremath{\mathrm{DEL}}}
\newcommand{\DELf}{\ensuremath{{\DEL}_{\!f}}}
\newcommand{\DELo}{\ensuremath{{\DEL}_{\omega}}}

\newcommand{\DL}{\ensuremath{\mathrm{DL}}}

\renewcommand{\H}{\ensuremath{\mathbf{H}}} 
\newcommand{\T}{\ensuremath{\mathbf{T}}} 
\newcommand{\M}{\ensuremath{\mathbf{M}}}

\newcommand{\DBox}[1]{\ensuremath{[#1]\,}}
\newcommand{\DDia}[1]{\ensuremath{\langle#1\rangle\,}}
\newcommand{\Rel}[2]{\ensuremath{{\parallel}{#1}{\parallel}^{#2}}}

\newcommand{\intervc}[2]{[#1..#2]}
\newcommand{\intervo}[2]{[#1..#2)}
\newcommand{\rangec}[3]{#1 \in \intervc{#2}{#3}}
\newcommand{\rangeo}[3]{#1 \in \intervo{#2}{#3}}

\newcommand{\Lab}[1]{\ensuremath{{\ell_{#1}}}}

\newcommand{\next}{{\ensuremath{\circ}}} 
\newcommand{\previous}{{\ensuremath{\bullet}}} 
\newcommand{\wnext}{\ensuremath{\widehat{\next}}}
\newcommand{\wprevious}{\ensuremath{\widehat{\previous}}}
\newcommand{\alwaysF}{\ensuremath{\square}}
\newcommand{\alwaysP}{\ensuremath{\blacksquare}}
\newcommand{\eventuallyF}{\ensuremath{\Diamond}}
\newcommand{\eventuallyP}{\ensuremath{\blacklozenge}}
\newcommand{\until}{\ensuremath{\mathbin{\bm{\mathsf{U}}}}}
\newcommand{\release}{\ensuremath{\mathbin{\bm{\mathsf{R}}}}}
\newcommand{\since}{\ensuremath{\mathbin{\bm{\mathsf{S}}}}}
\newcommand{\trigger}{\ensuremath{\mathbin{\bm{\mathsf{T}}}}}
\newcommand{\finally}{\ensuremath{\bm{\mathsf{F}}}}
\newcommand{\initially}{\ensuremath{\bm{\mathsf{I}}}}

\newcommand{\trivaluation}[3]{\ensuremath{\bm{#3}(#1,#2)}}
\newcommand{\trival}[2]{\trivaluation{#1}{#2}{m}} 
\newcommand{\trivalp}[2]{{\bm{m'}}(#1,#2)} 

\newcommand{\sysfont}{\textit}

\newcommand{\clingo}{\sysfont{clingo}}
\newcommand{\telingo}{\sysfont{telingo}}

\def\imp{\text{imp}}
\newcommand{\qed}{QED}

\def\stp{\uptau}

\def\FL{\mathit{FL}} 

\begin{document}

\title{Implementing Dynamic Answer Set Programming\\ over finite traces}

\author{%
  Pedro Cabalar\institute{University of Corunna, Spain}
  \and
  Mart\'{\i}n Di\'eguez\institute{University of Pau, France}
  \and
  Torsten Schaub\institute{University of Potsdam, Germany}
  \and
  Francois Laferriere$^3$
}

\maketitle

\begin{abstract}
We introduce an implementation of an extension of Answer Set Programming (ASP) with language constructs from dynamic (and temporal) logic
that provides an expressive computational framework for modeling dynamic applications.
Starting from logical foundations, provided by dynamic and temporal equilibrium logics over finite linear traces,
we develop a translation of dynamic formulas into temporal logic programs.
This provides us with a normal form result establishing the strong equivalence of formulas in different logics.
Our translation relies on the introduction of auxiliary atoms to guarantee polynomial space complexity
and to provide an embedding that is doomed to be impossible over the same language.
Finally,
the reduction of dynamic formulas to temporal logic programs allows us to extend ASP with both approaches in a uniform way
and to implement both extensions via temporal ASP solvers such as \telingo.
\end{abstract}
%

\section{Introduction}\label{sec:introduction}

Humans are not bothered at all when it comes to choosing a way home among a plethora of alternatives.
Similarly, in reasoning about action or planning, the underlying specifications admit an abundance of feasible plans.
Among them, we usually find only a few reasonable alternatives while the majority bear an increasing number of redundancies,
leading to infinite solutions in the worst case.
Popular ways to counterbalance this are to extend the specification by objectives, like shortest plans,
and/or imposing limits on the plan length.
Both are often implemented with optimization procedures and/or incremental reasoning methods (extending insufficient plan lengths).
However, computing a valid plan with such techniques usually involves solving numerous sub-problems
of nearly the same scale as the actual problem --- not to mention that solving somehow optimal plans is often particularly hard
(due to phase transition phenomena).
This rules out such techniques when it comes to highly demanding dynamic problems,
as we witnessed in applications to robotic intra-logistics~\cite{geobotscsangso18a}.

This motivates our approach to pair action theories with control theories in order to restrict our attention to plans selected by the control theory
among all feasible ones induced by the action theory.
This approach was pioneered by Levesque et~al.~in~\cite{lereleli97a} by combining
action theories in the Situation Calculus with
control programs expressed in Golog.
The rough idea is that a plan induced by an action theory must be compatible with a run of the associated Golog program.
Although Golog's design was inspired by Dynamic Logic~\cite{hatiko00a},
its semantics is given by a reduction to first-order logic.
Unlike this, we developed in~\cite{bocadisc18a,cadisc19a} the foundations of an approach integrating Answer Set Programming
(ASP~\cite{lifschitz19a}) with (linear) Dynamic Logic.
More precisely, we are interested in the combination of the logic of Here-and-There (\HT~\cite{heyting30a})
with linear Dynamic Logic over finite traces~(\LDLf~\cite{giavar13a}), called
Dynamic logic of Here-and-There (\DHTf) and particularly its non-monotonic extension, Dynamic Equilibrium Logic (\DELf~\cite{cadisc19a}).
We review both logics in Section~\ref{sec:del}.
ASP as such is known to constitute a fragment of Equilibrium Logic~\cite{pearce06a}.

In what follows,
our focus lies on implementing temporal and dynamic ASP via a reduction to regular ASP.
This aligns with the above discussion insofar as temporal logic programs, featuring one step operators,
are well-suited for providing action theories, while dynamic formulas allow for imposing compatibilities with path expressions,
that amount to regular expressions over primitive actions.%
\footnote{Temporal formulas constitute a proper fragment of \DHTf\ (cf.~Section~\ref{sec:del}).}
To this end, in this paper, we present the following contributions:
\begin{enumerate}[(i)]
\item \label{it1} we develop a three-valued characterization of \DHTf;
\item \label{it2} we establish a normal form for \DHTf\ showing that dynamic formulas can be reduced to (so-called) temporal logic programs;
\item \label{it3} we use this reduction to implement a solver accepting dynamic formulas in \DELf.
\end{enumerate}
For~\eqref{it1}, we extend the three-valued characterization from~\cite{goedel32a} to dynamic formulas,
something that allows for greatly simplifying proofs and has thus benefits well beyond this paper.
Also, it is, to the best of our knowledge, the first time this type of construction is used to capture dynamic logics and thus path expressions.
This three-valued definition also allows for establishing~\eqref{it2},
which shows that any dynamic formula can be equivalently reduced to a syntactic fragment called temporal logic programs.
This translation relies on the introduction of auxiliary atoms (in a Tseitin-style~\cite{tseitin68a}) for,
first, guaranteeing that its result is of polynomial size wrt the input formula,
and, second, surmounting the fact that translations of dynamic into temporal formulas are usually impossible without extending the language.
We explain both issues in more detail in Section~\ref{sec:tlp}.
Finally, the great benefit of this reduction is that we can use it for~\eqref{it3}, that is, implementing dynamic formulas in \DELf,
since temporal logic programs can be processed by an existing solver for temporal ASP, viz.\ \telingo~\cite{cakamosc19a}.
We describe the resulting implementation in Section~\ref{sec:system} and show the potential impact of pairing action and control theories
via an empirical analysis of an elevator scenario borrowed from~\cite{lereleli97a}.


%
\section{Linear Dynamic Equilibrium Logic}
\label{sec:del}

We start from the syntax of \emph{Linear Dynamic Logic} (\LDL) defined in~\cite{giavar13a}.
Given a set \PV{} of propositional variables (called \emph{alphabet}),
\emph{dynamic formulas} $\varphi$ and \emph{path expressions} $\rho$ are mutually defined by the pair of grammar rules:
\begin{align*}
  \varphi &::= \myatom \mid \bot \mid \top \mid \; \DBox{\rho} \varphi \; \mid \; \DDia{\rho} \varphi \\
  \rho    &::=  \stp \mid \varphi ? \mid \rho + \rho \mid \rho\mathrel{;}\rho \mid \rho^{\ast} \mid \rho^-
\end{align*}
This syntax is similar to the one of Dynamic Logic (\DL~\cite{hatiko00a})
but differs in the construction of atomic path expressions:
while \DL\ uses a different alphabet for \emph{atomic actions},
in \LDL\ there is a unique alphabet \PV\ (atomic propositions) and
the only atomic path expression is the constant $\stp \not\in \PV$ (read as ``step'')
that we also write as $\top$ (see below), overloading the constant truth symbol.
As we show further below, the above language allows us to capture several derived operators, like the Boolean and temporal ones:
\[
  \arraycolsep=2pt\def\arraystretch{1.3}
  \begin{array}{rclp{1pt}rcl}
    \varphi \wedge \psi & \eqdef & \DDia{\varphi?} \psi          && \varphi \vee \psi & \eqdef & \DDia{\varphi?+\psi?} \top
    \\
    \varphi \to \psi & \eqdef & \DBox{\varphi?} \psi             && \neg \varphi & \eqdef & \varphi \to \bot
    \\
    \finally & \eqdef & [\top]\bot                               && \initially & \eqdef & [\top^-]\bot
    \\
    \next \varphi & \eqdef & \DDia{\top} \varphi                 && \previous \varphi & \eqdef & \DDia{\top^-} \varphi
    \\
    \wnext \varphi & \eqdef & \DBox{\top} \varphi                && \wprevious \varphi & \eqdef & \DBox{\top^-} \varphi
    \\
    \eventuallyF \varphi & \eqdef & \DDia{\top^*} \varphi        && \eventuallyP \varphi & \eqdef & \DDia{\top^{*-}} \varphi
    \\
    \alwaysF \varphi & \eqdef & \DBox{\top^*} \varphi            && \alwaysP \varphi & \eqdef & \DBox{\top^{*-}} \varphi
    \\
    \varphi \until \psi & \eqdef & \DDia{(\varphi?;\top)^*} \psi && \varphi \since \psi & \eqdef & \DDia{(\varphi?;\top)^{*-}} \psi
    \\
    \varphi \release \psi & \eqdef & (\psi \until (\varphi \wedge \psi)) \vee \alwaysF \psi
                                                                 &&
    \varphi \trigger \psi & \eqdef & (\psi \since (\varphi \wedge \psi)) \vee \alwaysP \psi
\end{array}
\]
All connectives are defined in terms of the dynamic operators \DDia{\cdot} and \DBox{\cdot}.
This involves the Booleans' $\wedge$, $\vee$, and $\to$,
among which the definition of $\to$ is most noteworthy since it hints at the implicative nature of \DBox{\cdot}.
Negation $\neg$ is then expressed via implication, as usual in \HT{}.
Then, \DDia{\cdot} and \DBox{\cdot} also allow defining the future temporal operators
\finally,
\next,
\wnext,
\eventuallyF,
\alwaysF,
\until,
\release,
standing for
\emph{final,
next,
weak next,
eventually,
always,
until,} and
\emph{release},
and their past-oriented counterparts:
\initially,
\previous,
\wprevious,
\eventuallyP,
\alwaysP,
\since,
\trigger.
The weak one-step operators, \wnext{} and \wprevious, are of particular interest when dealing with finite traces,
since their behavior differs from their genuine counterparts only at the ends of a trace.
In fact, $\wnext \varphi$ can also be expressed as $\next \varphi \vee \finally$ (and $\wprevious$ as $\previous \varphi \vee \initially$).
%
%
A  formula is \emph{propositional}, if all its connectives are Boolean,
and \emph{temporal}, if it includes only Boolean and temporal ones.
%
%
%
As usual, a \emph{(dynamic) theory} is a set of (dynamic) formulas.
Following the definition of \emph{linear}~\DL{} (\LDL) in~\cite{giavar13a},
we sometimes use a propositional formula $\phi$ as a path expression actually standing for $(\phi?;\stp)$.
This means that the reading of $\top$ as a path expression amounts to $(\top?;\stp)$
which is just equivalent to $\stp$, as we see below.
Another abbreviation is the sequence of $n$ repetitions of some expression $\rho$ defined as $\rho^0 \eqdef \top?$ and $\rho^{n+1} \eqdef \rho; \rho^n$.
For instance, $\rho^3=\rho;\rho;\rho;\top?$ which amounts to $\rho;\rho;\rho$, as we see below.

Given $a \in \mathbb{N}$ and $b \in \mathbb{N} \cup \{\omega\}$,
we let $\intervc{a}{b}$ stand for the set $\{i \in \mathbb{N} \mid a \leq i \leq b\}$ and
$\intervo{a}{b}$ for $\{i \in \mathbb{N} \mid a \leq i < b\}$.
For the semantics, we start by defining a \emph{trace} of length $\lambda$ over alphabet \PV{} as a sequence $\tuple{H_i}_{\rangeo{i}{0}{\lambda}}$ of sets $H_i\subseteq\PV$.
A trace is \emph{infinite} if $\lambda=\omega$ and \emph{finite} otherwise, that is, $\lambda=n$ for some natural number $n \in \mathbb{N}$.
Given traces $\H=\tuple{H_i}_{\rangeo{i}{0}{\lambda}}$ and $\H'=\tuple{H'_i}_{\rangeo{i}{0}{\lambda}}$ both of length $\lambda$, we write $\H\leq\mathbf\H'$ if $H_i\subseteq H'_i$ for each $\rangeo{i}{0}{\lambda}$;
accordingly, $\mathbf{H}<\mathbf{H'}$ iff both $\mathbf{H}\leq\mathbf{H'}$ and $\mathbf{H}\neq\mathbf{H'}$.

Although \DHT\ shares the same syntax as \LDL, its semantics relies on traces whose states are pairs of sets of atoms.
A \emph{Here-and-There trace} (for short \emph{\HT-trace}) of length $\lambda$ over alphabet \PV{} is a sequence of pairs $\tuple{H_i,T_i}_{\rangeo{i}{0}{\lambda}}$ such that $H_i\subseteq T_i\subseteq \PV$ for any $\rangeo{i}{0}{\lambda}$.
As before, an \HT-trace is infinite if $\lambda=\omega$ and finite otherwise.
The intuition of using these two sets stems from \HT\ and Equilibrium Logic:
atoms in $H_i$ are those that can be proved;
atoms not in $T_i$ are those for which there is no proof;
and, finally, atoms in $T_i\setminus H_i$ are assumed to hold, but have not been proved.
We often represent an \HT-trace as a pair of traces $\tuple{\H,\T}$ of length $\lambda$
where $\H=\tuple{H_i}_{\rangeo{i}{0}{\lambda}}$ and $\T=\tuple{T_i}_{\rangeo{i}{0}{\lambda}}$ and $\H \leq \T$.
The particular type of \HT-traces that satisfy $\H=\T$ are called \emph{total}.
Given any \HT-trace $\M=\tuple{\H,\T}$,
we define \DHT\ satisfaction of formulas, namely, $\M,k \models \varphi$,
in terms of an accessibility relation for path expressions $\Rel{\rho}{\M} \subseteq \mathbb{N}^2$
whose extent depends again on $\models$ by double, structural induction.
%
\begin{definition}[\DHT{} satisfaction~\cite{cadisc19a}]\label{def:dht:satisfaction}
  An \HT-trace $\M=\tuple{\H,\T}$ of length $\lambda$ over alphabet \PV{}
  \emph{satisfies} a dynamic formula $\varphi$ at time point $\rangeo{k}{0}{\lambda}$,
  written \mbox{$\M,k \models \varphi$}, if the following conditions hold:
  \begin{enumerate}
  \item $\M,k \models \top$ and  $\M,k \not\models \bot$
  \item $\M,k \models \myatom$ if $\myatom \in H_k$ for any atom $\myatom \in \PV$
  \item \label{def:dhtsat.2} $\M, k \models \DDia{\rho} \varphi$
    if $\M,i \models \varphi$
    for some $i$ with $(k,i) \in \Rel{\rho}{\M}$
  \item \label{def:dhtsat.3} $\M, k \models \DBox{\rho} \varphi$
    if $\M',i \models \varphi$
    for all $i$ with $(k,i) \in \Rel{\rho}{\M'}$ \\ for both $\M'=\M$ and $\M'=\tuple{\T,\T}$
  \end{enumerate}
  where, for any \HT-trace $\M$, $\Rel{\rho}{\M} \subseteq \mathbb{N}^2$ is a relation on pairs of time points inductively defined as follows.
  \begin{enumerate}
  \setcounter{enumi}{4}
  \item $\Rel{\stp}{\M}            \ \eqdef\ \{ (k,k+1) \ \mid \rangeo{k,k+1}{0}{\lambda} \}$
  \item $\Rel{\varphi?}{\M}        \ \eqdef\ \{ (k,k) \mid  \M,k \models \varphi \}$
  \item $\Rel{\rho_1\mathrel{+}\rho_2}{\M} \ \eqdef\ \Rel{\rho_2}{\M} \cup \Rel{\rho_2}{\M}$
  \item
  $\!\!\!
  \begin{array}[t]{r@{\,}c@{\,}l}
    \Rel{\rho_1\mathrel{;}\rho_2}{\M}  \eqdef  \{ \ (k,i)  & | & {(k,j) \in \Rel{\rho_1}{\M}} \text{ and } \\
                                                           &   & {\,(j,i)\kern1pt\in \Rel{\rho_2}{\M}} \text{ for some } k \ \}
  \end{array}
  $
  \item $\Rel{\rho^*}{\M}          \ \eqdef\  \bigcup_{n\geq 0} \Rel{\rho^n}{\M}$
  \item $\Rel{\rho^-}{\M}          \ \eqdef\ \{ (k,i)\mid (i,k)\in\Rel{\rho}{\M} \}$
  \end{enumerate}
\end{definition}
%
We see that $\DDia{\rho}{\varphi}$ and $\DBox{\rho}{\varphi}$ quantify over time points $i$
that are reachable via path expression $\rho$ from the current point $k$,
that is, $(k,i) \in \Rel{\rho}{\M} \subseteq \intervo{0}{\lambda}\times\intervo{0}{\lambda}$.
An \HT-trace $\M$ is a \emph{model} of a dynamic theory $\Gamma$ if $\M,0 \models \varphi$ for all $\varphi \in \Gamma$.
We write $\DHT(\Gamma,\lambda)$ to stand for the set of \DHT{} models of length $\lambda$ of a theory $\Gamma$,
and define $\DHT(\Gamma) \eqdef \bigcup_{\lambda=0}^\omega \DHT(\Gamma,\lambda)$, that is, the whole set of models of $\Gamma$ of any length.
When $\Gamma=\{\varphi\}$ we just write $\DHT(\varphi,\lambda)$ and $\DHT(\varphi)$.

A formula $\varphi$ is a \emph{tautology} (or is \emph{valid}), written $\models \varphi$,
iff $\M,k \models \varphi$ for any \HT-trace \M\ and any $\rangeo{k}{0}{\lambda}$.
We call the logic induced by the set of all tautologies \emph{(Linear) Dynamic logic of Here-and-There} (\DHT{} for short).
Two formulas $\varphi, \psi$ are said to be \emph{equivalent}, written $\varphi \equiv \psi$,
whenever $\M,k \models \varphi$ iff $\M,k \models \psi$ for any \HT-trace $\M$ and any $\rangeo{k}{0}{\lambda}$.
This allows us to replace $\varphi$ by $\psi$ and vice versa in any context, and is the same as requiring that $\varphi \leftrightarrow \psi$ is a tautology.
Note that this relation, $\varphi \equiv \psi$, is stronger than coincidence of models, viz.\ $\DHT(\varphi)=\DHT(\psi)$.
For instance, $\DHT(\previous \top)=\DHT(\DDia{\top^-} \top)=\emptyset$
because models are checked at the initial situation $k=0$ and there is no previous situation at that point, so $\DHT(\previous \top)=\DHT(\bot)$.
However, in general, $\previous \top \not\equiv \bot$ since $\previous \top$ is satisfied for any $k>0$
(for instance $\next \previous \top \not\equiv \next \bot$ but $\next \previous \top \equiv \top$ instead).
As with formulas, we say that path expressions $\rho_1, \rho_2$ are \emph{equivalent},
written $\rho_1 = \rho_2$, whenever \mbox{$\Rel{\rho_1}{\M}=\Rel{\rho_2}{\M}$} for any \HT-trace $\M$.

The following equivalences of path expressions allow us to push the converse operator inside,
until it is only applied to $\stp$.
\begin{proposition}[\cite{cadisc19a}]\label{prop:eq:converse}
  For all path expressions $\rho_1$, $\rho_2$ and $\rho$ and for all formulas $\varphi$, the following equivalences hold:
  \begin{align*}
    (\rho^-)^- & =  \rho & (\varphi?)^- & =   \varphi? & (\rho^*)^- & =  (\rho^-)^*\\
    (\rho_1+ \rho_2)^- & =  \rho_1^- + \rho_2^- & (\rho_1; \rho_2)^- & =  \rho_2^- ; \rho_1^-
  \end{align*}
\end{proposition}
%
We say that $\varphi$ is in \emph{converse normal form} if all occurrences of the converse operator in $\varphi$ are applied to $\stp$.

We now introduce non-monotonicity by selecting a particular set of traces that we call \emph{temporal equilibrium models}.
First, given an arbitrary set $\mathfrak{S}$ of \HT{}-traces, we define the ones in equilibrium as follows.
%
\begin{definition}[Temporal Equilibrium/Stable models~\cite{cadisc19a}]\label{def:tem}
Let $\mathfrak{S}$ be some set of \HT-traces.
A total \HT-trace $\tuple{\T,\T} \in\mathfrak{S}$ is an \emph{equilibrium trace} of $\mathfrak{S}$ iff
there is no other $\tuple{\H,\T} \in\mathfrak{S}$ such that $\H < \T$.
\end{definition}
%
If $\tuple{\T,\T}$ is such an equilibrium trace, we also say that trace \T{} is a \emph{stable trace} of $\mathfrak{S}$.
We further talk about \emph{temporal equilibrium} or \emph{temporal stable models} of a theory $\Gamma$
when $\mathfrak{S}=\DHT(\Gamma)$, respectively.

We write $\DEL(\Gamma,\lambda)$ and $\DEL(\Gamma)$ to stand for the temporal equilibrium models of $\DHT(\Gamma,\lambda)$ and $\DHT(\Gamma)$ respectively.
Note that stable traces in $\DEL(\Gamma)$ are also \LDL{}-models of $\Gamma$ and, thus, \DEL{} is stronger than \LDL{}.
Besides, as the ordering relation among traces is only defined for a fixed $\lambda$,
it is easy to see the following result:
\begin{proposition}[\cite{cadisc19a}]
The set of temporal equilibrium models of $\Gamma$ can be partitioned by the trace length $\lambda$, that is,
$\bigcup_{\lambda=0}^\omega \DEL(\Gamma,\lambda) = \DEL(\Gamma)$.
\end{proposition}

(Linear) \emph{Dynamic Equilibrium Logic} (\DEL) is the non-monotonic logic induced by temporal equilibrium models of dynamic theories.
We obtain the variants \DELo{} and \DELf{} by applying the corresponding restriction to infinite or finite traces, respectively.

To illustrate non-monotonicity,
take the formula:
\begin{eqnarray}
 \DBox{(\neg h)^*} (\neg h \to s) \label{f:help}
\end{eqnarray}
whose reading is ``keep sending an sos ($s$) while no help ($h$) is perceived.''
Intuitively, $\DBox{(\neg h)^*}$ behaves as a conditional referring to any future state after $n \geq 0$ repetitions of $(\neg h?;\top)$.
Then, $\neg h \to s$ checks whether $h$ fails one more time at $k=n$: if so, it makes $s$ true again.
Without additional information, this formula has a unique temporal stable model per each length $\lambda$ satisfying $\Box (\neg h \wedge s)$, that is, $h$ is never concluded, and so, we repeat $s$ all over the trace.
Suppose we add now the formula $\DDia{\top^5} h$, that is, $h$ becomes true after five transitions.
Then, there is a unique temporal stable model for each $\lambda>5$ satisfying:
\[
\DDia{(\neg h \wedge s)^5; h \wedge \neg s; (\neg h \wedge \neg s)^*} \top
\]
Clearly, $\Box (\neg h \wedge s)$ is not entailed any more (under temporal equilibrium models) showing that \DEL{} is non-monotonic.
%


To conclude this section, we provide an alternative three-valued characterization of \DHT{} that is particularly useful for formal elaborations involving auxiliary atoms.
This alternative characterization relies on the idea of temporal three-valued interpretation in~\cite{cabalar10a} for the case of \TEL{} and is inspired, in its turn, in the characterization of HT in terms of G\"odel's $G_3$ logic~\cite{goedel32a}.
Under this orientation, we deal with three truth values $\{0,1,2\}$ standing for: $2$ (or proved true) meaning satisfaction ``here''; $0$ (or assumed false) meaning falsity ``there''; and $1$ (potentially true) that are formulas assumed true but not proved.
Given an \HT-trace $\M=\tuple{\H,\T}$ we define its associated \emph{truth valuations} as a pair of mutually recursive functions $\trival{k}{\varphi}$ and $\trival{k,i}{\rho}$ that assign a truth value in the set $\{0,1,2\}$ to formula $\varphi$ at time point $\rangeo{k}{0}{\lambda}$ or to the pair $(k,i)$ for path expression $\rho$, respectively.
The valuation of formulas follows the next rules:
\begin{eqnarray*}
	\trival{k}{\bot} & \eqdef & 0 \\
	\trival{k}{\top} & \eqdef & 2 \\
	\trival{k}{\myatom} & \eqdef &
	\begin{cases}
		0 & \text{if} \ \myatom \not\in T_k \\
		1 & \text{if} \ \myatom \in T_k\setminus H_k \\
		2 & \text{if} \ \myatom \in H_k
	\end{cases}\hspace{15pt} \text{for any atom } \myatom \in \PV\\
	\trival{k}{\DBox{\rho}\psi} & \eqdef & \min \ \{ \ \imp( \trival{k,i}{\rho} , \trival{i}{\psi}) \mid \rangeo{i}{0}{\lambda} \ \}\\&& \text{where }\\
	 & & \imp( x , y) \eqdef
	\begin{cases}
	2  & \text{ if } x \le y \\
	y  & \text{ otherwise } \\
      \end{cases}\\
  \trival{k}{\DDia{\rho}\psi} &\eqdef& \max \ \{ \ \min (\trival{k,i}{\rho},\trival{i}{\psi}) \mid \rangeo{i}{0}{\lambda} \ \}
\end{eqnarray*}
whereas the function for path expressions is defined as follows:
\begin{eqnarray*}
\trival{k,j}{\stp} & \eqdef &  \begin{cases}
                                2 & \hbox{ if } j = k+1\\
                                0 & \hbox{otherwise}.
                               \end{cases}\\
\trival{k,j}{\varphi?} &\eqdef&	\begin{cases}
                                 \trival{k}{\varphi} & \hbox{ if } j = k\\
                                 0 & \hbox{otherwise}.
                                \end{cases}	\\
\trival{k,j}{\rho_1\mathrel{+}\rho_2} &\eqdef& \max(\trival{k,j}{\rho_1},\trival{k,j}{\rho_2} )\\
\trival{k,j}{\rho_1\mathrel{;}\rho_2} &\eqdef& \max \lbrace \min(\trival{k,i}{\rho_1},\\&&\kern42pt\trival{i,j}{\rho_2}) \mid \rangeo{i}{0}{\lambda}  \;\rbrace\\
\trival{k,j}{\rho^*} &\eqdef& \max \{ \trival{k,j}{\rho^n} \mid \text{for all } n\geq0 \}\\
\trival{k,j}{\rho^-} & \eqdef  &  \trival{j,k}{\rho}
\end{eqnarray*}

This results in the following three-valued characterisation of \HT-traces in \DELf.
\footnote{An extended version of the paper with the proofs is available at
  \url{http://arxiv.org/abs/2002.06916}}
\begin{theorem}\label{prop:three-valued}
  Let $\tuple{\H,\T}$ be a \HT-trace of length $\lambda$,
  ${\bm m}$ its associated valuation and $\rangeo{k}{0}{\lambda}$:
  \begin{enumerate}
  \item $\tuple{\H,\T},k \models \varphi$ iff $\trival{k}{\varphi}=2$
  \item $\tuple{\T,\T},k \models \varphi$ iff $\trival{k}{\varphi}\neq 0$
    \smallskip
  \item $(k,j) \in \Rel{\rho}{\tuple{\H,\T}}$ iff $\trival{k,j}{\rho} = 2$
  \item $(k,j) \in \Rel{\rho}{\tuple{\T,\T}}$ iff $\trival{k,j}{\rho} \neq 0$
  \end{enumerate}
\end{theorem}
%


\section{Reduction to temporal logic programs}
\label{sec:tlp}

In this section, we elaborate upon a reduction of arbitrary dynamic formulas%
\footnote{This covers finite dynamic theories, understood as the conjunction of their formulas.}
into a syntactic subclass called \emph{temporal logic programs}~\cite{cakascsc18a}.
A temporal logic program is a conjunction of temporal formulas with a restricted syntax that, when interpreted under temporal stable models, have a close relation to rules from disjunctive logic programming.
Temporal logic programs were proved in~\cite{cakascsc18a} to constitute a normal form for \TELf\
(if we allow for auxiliary atoms) and
used later on as a basic syntax for the temporal ASP system \telingo~\cite{cakamosc19a}.
We proceed next to describe their syntax.

Given a set of propositional variables $\PV$,
we define the set of \emph{temporal literals} as
\(
\lbrace a, \neg a,  \previous a, \neg \previous a, \mid a \in \PV\rbrace.
\)
A temporal logic program is a set formed by three different types of rules:
\begin{enumerate}
\item an \emph{initial rule} is of the form $B \rightarrow A$
\item a \emph{dynamic rule} is of the form $\wnext\alwaysF\left(B\rightarrow A\right)$
\item a \emph{final rule} is of the form $\alwaysF\left(\finally \rightarrow \left(B \rightarrow A\right)\right)$
\end{enumerate}
where $B = b_1 \wedge \cdots \wedge b_n$ (with $n\ge 0$) and $A = a_1 \vee \cdots \vee a_m$ (with $m \ge 0$)
and $b_i$ and $a_j$ are temporal literals in the case of dynamic rules and regular literals
(i.e.\ $\lbrace a , \neg a \mid a \in \PV \rbrace$)
in the case of initial and final rules.
We also allow for \emph{global rules} of the form $\alwaysF\left(B\rightarrow A\right)$
that stand for the conjunction of an initial rule $B \to A$ and a dynamic rule $\wnext\alwaysF\left(B\rightarrow A\right)$.
\begin{theorem}[Normal form]\label{thm:nf}
Every dynamic formula $\gamma$ can be converted into a temporal program being \DHTf{}-equivalent to $\gamma$.
\end{theorem}
To prove the above theorem,
we provide a sound transformation from any dynamic formula $\gamma$ into a temporal logic program $\pi(\gamma)$.
As a first step, we assume that $\gamma$ is already in converse normal form: this can be achieved by repeatedly applying the equivalences in Proposition~\ref{prop:eq:converse}.
The reduction of $\gamma$ into temporal program $\pi(\gamma)$ uses an extended alphabet $\mathcal{A}^+\supseteq\mathcal{A}$ that additionally contains new atoms \Lab{\varphi} (aka label) for formulas $\varphi$ over $\mathcal{A}$ that are either subformulas of $\gamma$ or elaborations of them.
This set of formulas is called the Fisher-Ladner closure~\cite{fislad79a} of $\gamma$ and formally defined below.
%
\begin{definition}[Fisher-Ladner closure~\cite{fislad79a}]
  The Fisher-Ladner closure $\FL(\gamma)$ of a dynamic formula $\gamma$  (in converse normal form) is a set of dynamic formulas inductively defined as follows:
  \begin{enumerate}
  \item $\gamma \in \FL(\gamma)$
  \item $(\varphi \otimes \psi) \in \FL(\gamma)$ implies $\varphi \in \FL(\gamma)$ and $\psi \in \FL(\gamma)$,
    \par where $\otimes \in \lbrace \wedge, \vee, \rightarrow\rbrace$
  \item If $\DDia{\rho}\varphi \in \FL(\gamma)$ then $\varphi \in \FL(\gamma)$
  \item If $\DBox{\rho}\varphi \in \FL(\gamma)$ then $\varphi \in \FL(\gamma)$
  \item If $\DDia{\psi?}\varphi \in \FL(\gamma)$ then $\psi \in \FL(\gamma)$ and $\varphi \in \FL(\gamma)$
  \item If $\DBox{\psi?}\varphi \in \FL(\gamma)$ then $\psi \in \FL(\gamma)$ and $\varphi \in \FL(\gamma)$
  \item If $\DDia{\rho_1\mathrel{;}\rho_2}\varphi \in \FL(\gamma)$ then $\DDia{\rho_1}\DDia{\rho_2}\varphi \in \FL(\gamma)$
  \item If $\DBox{\rho_1\mathrel{;}\rho_2}\varphi \in \FL(\gamma)$ then $\DBox{\rho_1}\DBox{\rho_2}\varphi \in \FL(\gamma)$
  \item If $\DDia{\rho_1\mathrel{+}\rho_2}\varphi \in \FL(\gamma)$ then $\DDia{\rho_1}\varphi \in \FL(\gamma)$ and $\DDia{\rho_2}\varphi \in \FL(\gamma)$
  \item If $\DBox{\rho_1\mathrel{+}\rho_2}\varphi \in \FL(\gamma)$ then $\DBox{\rho_1}\varphi \in \FL(\gamma)$ and $\DBox{\rho_2}\varphi \in \FL(\gamma)$
  \item If $\DDia{\rho^*}\varphi\in \FL(\gamma)$ then $\DDia{\rho}\DDia{\rho^*}\varphi \in \FL(\gamma)$
  \item If $\DBox{\rho^*}\varphi\in \FL(\gamma)$ then $\DBox{\rho}\DDia{\rho^*}\varphi \in \FL(\gamma)$
  \end{enumerate}
  Any set satisfying these conditions is called \emph{closed}.
\end{definition}
\begin{proposition}
For any dynamic formula $\gamma$, its closure $\FL(\gamma)$ is finite.
\end{proposition}
%
Thus, given the dynamic formula $\gamma$ on alphabet $\mathcal{A}$ to be translated,
we define the extended alphabet $\mathcal{A}^+ \eqdef \mathcal{A} \cup \{\Lab{\mu} \mid \mu \in \FL(\gamma)\}$.
For convenience, we simply use $\Lab{\varphi} \overset{\mathit{def}}{=} \varphi $ if $\varphi$ is $\top , \bot$ or an atom $a \in \mathcal{A}$.

As happened with the normal form reduction for \TELf\ in~\cite{cakascsc18a}, the translation is done in two phases: we first obtain a temporal theory containing double implications, and then we unfold them into  temporal rules.
We start by defining the temporal theory $\sigma(\gamma)$ that introduces new labels $\Lab{\mu}$ for each formula $\mu \in \FL(\gamma)$.
This theory contains the formula $\Lab{\gamma}$ and, per each label $\Lab{\mu}$,
a set of formulas $\eta(\mu)$ fixing the label's truth value.
Formally:
\begin{align*}
\sigma(\gamma) =
\left\lbrace \Lab{\gamma} \right\rbrace
\cup \left\lbrace \eta(\mu)
\mid \mu \in \mathit{FL}(\Gamma)\right\rbrace
\end{align*}
Table~\ref{tab:translation} shows the definitions $\eta(\mu)$ for each $\mu$ in the closure $\FL(\gamma)$ depending on the outer modality in the formula.
\begin{table}
\centering
\[
\renewcommand{\arraystretch}{1.5}
\begin{array}{|c|l|}

\hline
\mu\in \FL(\gamma) & \eta(\mu) \\ \hline
\DDia{\stp} \varphi & \wnext \alwaysF (\previous \Lab{\mu} \leftrightarrow \Lab{\varphi})
\hspace{20pt}  \alwaysF (\finally \to \neg \Lab{\mu}) \\ \hline
\DBox{\stp} \varphi & \wnext \alwaysF (\previous \Lab{\mu} \leftrightarrow \Lab{\varphi})
\hspace{20pt}
\alwaysF (\finally \to \Lab{\mu}) \\ \hline
\DDia{\stp^-} \varphi & \wnext \alwaysF ( \Lab{\mu} \leftrightarrow \previous \Lab{\varphi})
\hspace{20pt}   \neg \Lab{\mu} \\ \hline
\DBox{\stp^-} \varphi & \wnext \alwaysF ( \Lab{\mu} \leftrightarrow \previous \Lab{\varphi})
\hspace{20pt}   \Lab{\mu} \\ \hline
\DDia{\varphi?} \psi & \alwaysF ( \Lab{\mu} \leftrightarrow \Lab{\varphi} \wedge \Lab{\psi}) \\ \hline
\DBox{\varphi?} \psi & \alwaysF ( \Lab{\mu} \leftrightarrow (\Lab{\varphi} \to \Lab{\psi})) \\ \hline
\DDia{\rho+\rho'} \varphi & \alwaysF ( \Lab{\mu} \leftrightarrow \Lab{\alpha} \vee \Lab{ \beta})
\hspace{20pt} \text{with } \alpha=\DDia{\rho} \varphi, \beta=\DDia{\rho'} \varphi\\ \hline
\DBox{\rho+\rho'} \varphi & \alwaysF ( \Lab{\mu} \leftrightarrow \Lab{\alpha} \wedge \Lab{ \beta})
\hspace{20pt} \text{with } \alpha=\DBox{\rho} \varphi, \beta=\DBox{\rho'} \varphi\\ \hline
\DDia{\rho;\rho'} \varphi & \eta(\ \DDia{\rho} \DDia{\rho'} \varphi\ ) \\ \hline
\DBox{\rho;\rho'} \varphi & \eta(\ \DBox{\rho} \DBox{\rho'} \varphi\ ) \\ \hline
\DDia{\rho^*} \varphi & \alwaysF (\Lab{\mu} \leftrightarrow \Lab{\varphi} \vee \Lab{\alpha})
\hspace{20pt} \text{with } \alpha=\DDia{\rho} \DDia{\rho^*} \varphi\\
 & \alwaysF (\finally \to (\Lab{\mu} \leftrightarrow \Lab{\varphi}) ) \\ \hline
\DBox{\rho^*} \varphi & \alwaysF (\Lab{\mu} \leftrightarrow \Lab{\varphi} \wedge \Lab{\alpha})
\hspace{20pt} \text{with } \alpha=\DBox{\rho} \DBox{\rho^*} \varphi\\
 & \alwaysF (\finally \to (\Lab{\mu} \leftrightarrow \Lab{\varphi}) ) \\ \hline
\end{array}
\]
\caption{Normal form translation}
\label{tab:translation}
\end{table}

As an example, take the dynamic formula $\gamma=\DDia{(p?;\top)^*} q$ (which corresponds to the temporal formula $p \until q$).
In the first step, we get
\begin{eqnarray}
\alwaysF (\Lab{\gamma} \leftrightarrow q \vee \Lab{\alpha}) \label{f:ex1.1} \\
\alwaysF (\finally \to (\Lab{\gamma} \leftrightarrow q) ) \label{f:ex1.2}
\end{eqnarray}
where we just used $q$ as its label $\Lab{q}$, and $\alpha$ stands for $\DDia{p?;\top} \DDia{(p?;\top)^*} q$ which belongs to $\FL(\gamma)$.
The truth of $\Lab{\alpha}$ is determined by $\eta(\alpha)$ which, in the table, is first unfolded into $\eta(\DDia{p?} \ \DDia{\top} \ \DDia{(p?;\top)^*} q)$ leading to:
\begin{eqnarray}
\alwaysF ( \Lab{\alpha} \leftrightarrow p \wedge \Lab{\beta})  \label{f:ex1.3}
\end{eqnarray}
with $\beta=\DDia{\top} \ \DDia{(p?;\top)^*} q$ also in $\FL(\gamma)$.
Notice now that $\beta$ contains $\gamma$ as a subformula, so it can be written as: $\beta=\DDia{\top} \ \gamma$.
Then $\eta(\beta)$ just corresponds to the pair of formulas:
\begin{eqnarray}
\wnext \alwaysF (\previous \Lab{\beta} \leftrightarrow \Lab{\gamma})  \label{f:ex1.4}\\
\alwaysF (\finally \to \neg \Lab{\beta})  \label{f:ex1.5}
\end{eqnarray}
and the whole translation amounts to $\sigma(\gamma)=\{\Lab{\gamma}\} \cup \{\eqref{f:ex1.1}-\eqref{f:ex1.5}\}$.

As a second example, consider the formula $\gamma=\DBox{(\top;\top)^*} p$ meaning that $p$ holds in all even time points (this formula is well-known not to be \LTL\ representable).
The formulas obtained for $\eta(\gamma)$ are:
\begin{eqnarray}
\alwaysF (\Lab{\gamma} \leftrightarrow p \wedge \Lab{\alpha})\\
 \alwaysF (\finally \to (\Lab{\gamma} \leftrightarrow p) )
\end{eqnarray}
with $\alpha=\DBox{\top;\top} \DBox{(\top;\top)^*} p$ and $\eta(\alpha)=\eta(\DBox{\top} \DBox{\top} \DBox{(\top;\top)^*)} p)$ is:
\begin{eqnarray}
\wnext \alwaysF (\previous \Lab{\alpha} \leftrightarrow \Lab{\beta}) \\
\alwaysF (\finally \to \Lab{\alpha}) \label{f:notcentered}
\end{eqnarray}
with $\beta= \DBox{\top} \DBox{(\top;\top)^*)} p$.
Since $\beta$ is actually $\DBox{\top} \gamma$, we get $\eta(\beta)$:
\begin{eqnarray}
\wnext \alwaysF (\previous \Lab{\beta} \leftrightarrow \Lab{\gamma}) \\
\alwaysF (\finally \to \Lab{\beta})
\end{eqnarray}

As we have seen, in the general case, formulas in $\eta(\mu)$ are not temporal rules, since they sometimes contain double implications.
However, they all have the forms $\varphi$, $\alwaysF \varphi$, $\wnext \alwaysF \varphi$ or $\alwaysF (\finally \to \varphi)$, for some inner propositional formula $\varphi$ formed with temporal literals.
For instance, in~\eqref{f:ex1.1}, the inner $\varphi$ corresponds to the propositional formula $\Lab{\gamma} \leftrightarrow p \vee \Lab{\alpha}$.
As first shown in~\cite{cabfer08a} propositional formulas in \HT{} can be reduced to conjunctions of disjunctive rules.
In this way, we can apply \HT{} transformations (as those in~\cite{capeva05a}) and THT axioms~\cite{baldie16} to eventually obtain a temporal logic program.
In the case of~\eqref{f:ex1.1}, the inner double implication is unfolded into three rules that, after applying property $\alwaysF (\alpha \wedge \beta) \leftrightarrow \alwaysF \alpha \wedge \alwaysF \beta$, eventually lead to:
\[
  \alwaysF(\Lab{\gamma} \to q \vee \Lab{\alpha}) \hspace{20pt}
  \alwaysF(q \to \Lab{\gamma}) \hspace{20pt}
  \alwaysF(\Lab{\alpha} \to \Lab{\gamma})
\]

Given an \HT-trace{}
\(
\tuple{\H,\T} \eqdef \tuple{H_i,T_i}_{i=0}^{\lambda}
\),
we define its restriction to alphabet $\mathcal{A}$
as
\(
\tuple{\H,\T}|_{\mathcal{A}}
\eqdef
\tuple{H_i\cap\mathcal{A},T_i\cap\mathcal{A}}_{i=0}^{\lambda}
\).
Similarly, for any set $S$ of \HT-traces we write $S|_{\mathcal{A}}$ to stand for
$\{ \tuple{\H,\T}|_{\mathcal{A}} \mid \tuple{\H,\T} \in S\}$ as expected.

The following lemma shows that $\Lab{\mu}$ and $\mu$ are equivalent:
\begin{lemma}\label{lem:nf2}
  Let $\gamma$ be a dynamic formula over $\mathcal{A}$ and
  let $\tuple{\H,\T}$ be a \DHTf{} model of $\sigma(\gamma)$ being associated with the three-valuation ${\bm m}$.

  Then, for any $\mu \in \mathit{FL}(\Gamma)$ and any $\rangeo{k}{0}{\lambda}$,
  we have $\trival{k}{\Lab{\mu}} = \trival{k}{\mu}$.
\end{lemma}
\begin{theorem}\label{lem:nf1}
  For any dynamic formula $\gamma$ and any length $\lambda$, we have
  $$
  \DHT(\gamma,\lambda) = \DHT(\sigma(\gamma),\lambda)|_{\mathcal{A}}.
  $$
\end{theorem}
\begin{corollary}
  Let $\gamma$ be a dynamic formula over $\mathcal{A}$.

  Then, translation $\sigma(\gamma)$ is \emph{strongly faithful}, that is:
  $$
  \DEL(\gamma \wedge \gamma') = \DEL(\sigma(\gamma)\wedge \gamma')|_{\mathcal{A}}.
  $$
  for any arbitrary dynamic formula $\gamma'$ over $\mathcal{A}$.
\end{corollary}
\begin{proposition}
  Translation $\sigma(\gamma)$ has a polynomial size with respect to the size of $\gamma$.
\end{proposition}

\section{Extending \telingo\ with dynamic formulas}
\label{sec:system}

We have extended the temporal ASP solver \telingo%
\footnote{Details on the functioning of \telingo\ are given in~\cite{cakamosc19a};
  its implementation is available at \url{https://github.com/potassco/telingo}.} with dynamic formulas over finite traces.
More precisely, the current version supports negated occurrences of dynamic formula, as in integrity constraints;
it is available at~\url{https://github.com/potassco/telingo/releases/tag/v2.0.0}.

To this end, \telingo\ provides (theory) atoms of form
`\lstinline[mathescape]+&del{$\varphi$}+'
that encapsulate arbitrary dynamic formulas $\varphi$.
Their syntax is given in Table~\ref{tab:operators};
it is supplied as a theory grammar to the underlying ASP system \clingo~\cite{gekakaosscwa16a}.%
\footnote{Note that \lstinline{#} and \lstinline{&} indicate basic ASP and customizable theory concepts.}
\begin{table*}
\[
\begin{array}{lp{1pt}c@{\quad}c@{\qquad}lp{20pt}c@{\quad}c@{\qquad}l}
        && {\texttt{\#true}}         & \top                      & \text{\emph{true}}              && {\texttt{\#false}}          & \bot                  & \text{\emph{false}}               \\[2pt]
  \TELf && {\texttt{\&initial}}      & {\initially}              & \text{\emph{initial}}           && {\texttt{\&final}}          & {\finally}            & \text{\emph{final}}               \\
        && {\texttt{'p}}             & {\previous p}             & \text{\emph{previous}}          && {\texttt{p'}}               & {\next p}             & \text{\emph{next}}                \\[2pt]
        && {\texttt{<}\varphi}       & {\previous\varphi}        & \text{\emph{previous}}          && {\texttt{>}\varphi}         & {\next\varphi}        & \text{\emph{next}}                \\
        && {\phi\texttt{<?}\varphi}  & {\phi\since\varphi}       & \text{\emph{since}}             && {\phi\texttt{>?}\varphi}    & {\phi\until\varphi}   & \text{\emph{until}}               \\
        && {\phi\texttt{<*}\varphi}  & {\phi\trigger\varphi}     & \text{\emph{trigger}}           && {\phi\texttt{>*}\varphi}    & {\phi\release\varphi} & \text{\emph{release}}             \\
        && {\texttt{<?}\varphi}      & {\eventuallyP\varphi}     & \text{\emph{eventually before}} && {\texttt{>?}\varphi}        & {\eventuallyF\varphi} & \text{\emph{eventually afterward}}\\
        && {\texttt{<*}\varphi}      & {\alwaysP\varphi}         & \text{\emph{always before}}     && {\texttt{>*}\varphi}        & {\alwaysF\varphi}     & \text{\emph{always afterward}}    \\
        && {\texttt{<:}\varphi}      & {\wprevious\varphi}       & \text{\emph{weak previous}}     && {\texttt{>:}\varphi}        & {\wnext\varphi}       & \text{\emph{weak next}}           \\[2pt]
 \DELf  &&                           &                           &                                 && {\rho \texttt{.>*} \varphi} & {\DBox{\rho}\varphi}  & \text{\emph{always}}              \\
        &&                           &                           &                                 && {\rho \texttt{.>?} \varphi} & {\DDia{\rho}\varphi}  & \text{\emph{eventually}}          \\[2pt]
        && {\texttt{?}\varphi}       & {\varphi?}                & \text{\emph{test}}              && {\texttt{*}\rho}            & {\rho^{\ast}}         & \text{\emph{star}}                \\
        && {\rho_1\texttt{;;}\rho_2} & {\rho_1\mathrel{;}\rho_2} & \text{\emph{sequence}}            && {\rho_1\texttt{+}\rho_2}    & {\rho_1+\rho_2}       & \text{\emph{choice}}
\end{array}
\]
\caption{Past and future temporal operators in \telingo{} and their \DELf{} and \TELf{} counterparts}
\label{tab:operators}
\end{table*}
%

%
The one of the dynamic operators {\DBox{\rho}} and {\DDia{\rho}} follows that of \alwaysF\ and \eventuallyF, respectively,
by extending `\lstinline{>*}' and `\lstinline{>?}' by prepending path expressions $\rho$ separated by a dot, viz.\
`\lstinline[mathescape]{$\rho\,$.$\,$>*}' and `\lstinline[mathescape]{$\rho\,$.$\,$>?}'\;.
For instance, the dynamic formula
\(
\DDia{(p?;\top)^*}q
\)
from Section~\ref{sec:tlp} is expressed as
\begin{lstlisting}[numbers=none]
    &del{ * (?p ;; &true ) . >? q }
\end{lstlisting}

The current restriction to negated dynamic formulas allows for an easier algorithmic treatment since formulas in the scope of negation are
interpreted as in \LDLf\ (just as negated formulas in \HT\ can be treated as in classical logic).
Although this restriction will be lifted in a next release,
it already supports an agreeable modeling methodology for dynamic domain separating action and control theories.
The idea is to model the actual action theory with temporal rules, fixing static and dynamic laws,
while the control theory, enforcing certain (sub)trajectories, is expressed by integrity constraints using dynamic formulas.
This is similar to the pairing of action theories in situation calculus and Golog programs~\cite{lereleli97a}.

Let us illustrate this with the example in Listing~\ref{lst:elevator} that aims at modeling a simple elevator.
This example is borrowed from~\cite{lereleli97a}.
%
\begin{lstlisting}[float,caption={\telingo\ encoding for the elevator problem},label=lst:elevator,basicstyle=\ttfamily\footnotesize]
#program always.

{wait; up; down; serve} = 1 :- not &final.
:- up,   at(X), not floor(X+1).
:- down, at(X), not floor(X-1).

at(X+1):- 'up,   'at(X).
at(X-1):- 'down, 'at(X).
at(X)  :- 'at(X), not 'up, not 'down.
called(X):- 'called(X), #false:'at(X), 'serve.

:- called(X), &final.

ready :- called(X), at(X).

#program initial.

:- not &del{ *( (*up + *down) ;; ?ready ;; serve)
	 ;; *wait .>? &final }.
\end{lstlisting}
%
The temporal program in Lines~1-14 constitutes the action theory;
it is expressed in \TELf.
All its rules in the scope of the program declaration headed by \lstinline{always} are thought of as being preceded by $\alwaysF$,
that is, they are added as global rules (see beginning of Section~\ref{sec:tlp}).
Line~3 tells us that exactly one of the four actions
$\mathit{wait}$,
$\mathit{up}$,
$\mathit{down}$,
or
$\mathit{serve}$,
occurs at any time before the end of the trace.
The next two lines check the preconditions of action $\mathit{up}$ and $\mathit{down}$.
Lines~7-9 provide effect and inertia axioms for the fluent $\mathit{at}$,
which reflects the current floor of the elevator.
Line~10 expresses that a call at a floor persists unless the floor was served.
Line~12 gives the actual goal condition,
requiring that no call remains unserved in the final state.
Line~14 indicates that the elevator is ready to serve,
whenever it is at a floor that it was called to.%
\footnote{The rule in Line~14 only provides an auxiliary atom used in Line~17 below,
  and may thus be regarded as not belonging the actual action theory.}

The integrity constraint in Line~18-19 represents the following dynamic formula:
\begin{gather}
  \label{eq:elevator:control}
  \bot\leftarrow\neg\DDia{((\mathit{up}^\ast+ \mathit{down}^\ast); \mathit{ready}?; \mathit{serve})^\ast; \mathit{wait}^\ast} \finally
\end{gather}
The purpose of this constraint is to eliminate fruitless wandering of the elevator.
More precisely,
it provides a simple control theory stipulating that
the elevator must pick one direction, either up or down,
and move in this direction until it reaches a floor to which it was called,
and serve this floor;
this process is repeated an arbitrary number of times;
finally, the elevator may have to wait until the end of the trace.
Note that~\eqref{eq:elevator:control} is posed as an initial temporal rule, as indicated by the program directive in Line~16.
Hence, its path expression must be matched by a trajectory from the initial to the final state,
enforced by \finally\ in~\eqref{eq:elevator:control} and
\lstinline{&final} in Line~19 in Listing~\ref{lst:elevator},
respectively.

For computation,
programs as in Listing~\ref{lst:elevator} are treated according to the translation introduced in Section~\ref{sec:tlp}.
That is, at the outset, all dynamic formulas are transformed into temporal rules.
Each dynamic formula $\gamma$ is recursively rewritten using translation $\eta$ until the formula is free of any dynamic constructs.
This is accomplished via \clingo's functionalities for manipulating abstract syntax trees.
Then, \telingo's API allows us to turn the obtained equivalences among temporal formulas directly into a regular logic program.
This involves the extension of predicates with time variables as well as the introduction of variables and rules reflecting
the successive lengthening of finite traces (cf.~\cite{cakamosc19a}; temporal rules are treated analogously).
This is needed to be able to solve temporal logic programs incrementally.
That is, traces of increasing length are investigated by incrementally extending the underlying logic program.
Once a model is found, the search stops and the corresponding traces are provided as output.
This amounts to computing a nonempty set $\DEL(P,\lambda)$ of stable traces for the smallest $\lambda\geq 0$
and some temporal program $P$ at hand.

Finally, let us examine the impact of the dynamic formula in Listing~\ref{lst:elevator}
on the trajectories induced by the action theory as well as solver performance.
Although we believe that~\eqref{eq:elevator:control} allows us to significantly reduce the number
of trajectories induced by the action theory in lines~1-14, its impact on search is less clear cut
because it comes with an augmentation of the number of constraints in the solver.

To analyze this, we consider the following simple elevator problems:
We look at $\texttt{n}=5,7,9,11$ floors, respectively, and
initially place a call at the ground and top floor while
the elevator sits on the middle floor.
The goal is to have all floors served at the end of a trace (cf.~Line~12).
This can be expressed by the following facts.
\begin{lstlisting}[numbers=none,basicstyle=\ttfamily\footnotesize]
#program always.   floor(1..n).

#program initial.  at((n+1)/2). called(1;n).
\end{lstlisting}

We run each instance with different horizons, beginning with the minimum lengths of satisfiable traces,
viz.~$\lfloor{\mathtt{3n}+1}\rfloor\mathtt{/2}$,
and gradually extending this by 1~to~4 to introduce more room for redundancies.
Our experiments were run with \telingo~2-$\alpha$ (based on \clingo~5.4.1) and
obtained without imposing any time or memory restrictions.
Our results are summarized in Table~\ref{tab:stats}.
%
\begin{table*}[t]
\newcommand{\MODELS}[2]{#1/#2}
\newcommand{\choices}[2]{#1/#2}
\newcommand{\conflicts}[2]{#1/#2}
\newcommand{\vars}[2]{#1/#2}
\newcommand{\constraints}[2]{#1/#2}
\newcommand{\resulta}[7]{#3}
\newcommand{\resultb}[7]{#4}
\newcommand{\resultc}[7]{#7}
\begin{center}
    \begin{tabular}{| c | r | r | r | r | r | l |}
        \hline

    $\lambda$ (horizon)& \multirow{2}{*}{$\lfloor{\mathtt{3n}+1}\rfloor\mathtt{/2}$} & \multirow{2}{*}{$\cdot+1$} &\multirow{2}{*}{$\cdot+2$}&  \multirow{2}{*}{$\cdot+3$} &\multirow{2}{*}{$\cdot+4$} & \multirow{2}{*}{indicators}\\ \cline{1-1}

    $\mathtt{n}$ (floors)  & & & & & & \\ \hline

\multirow{3}{*}{5} &
\resulta{5}{8}{\MODELS{2}{2}}{\choices{141}{2}}{\conflicts{92}{1}}{\vars{407}{821}}{\constraints{1119}{1929}} &
\resulta{5}{9}{\MODELS{34}{2}}{\choices{295}{7}}{\conflicts{184}{5}}{\vars{494}{955}}{\constraints{1402}{2306}} &
\resulta{5}{10}{\MODELS{340}{2}}{\choices{660}{7}}{\conflicts{248}{4}}{\vars{589}{1097}}{\constraints{1717}{2715}} &
\resulta{5}{11}{\MODELS{2618}{2}}{\choices{3183}{8}}{\conflicts{437}{4}}{\vars{692}{1247}}{\constraints{2064}{3156}} &
\resulta{5}{12}{\MODELS{17204}{2}}{\choices{19209}{11}}{\conflicts{1791}{4}}{\vars{803}{1405}}{\constraints{2443}{3629}} &
models \\ \cline{2-7}

&
\resultb{5}{8}{\MODELS{2}{2}}{\choices{141}{2}}{\conflicts{92}{1}}{\vars{407}{821}}{\constraints{1119}{1929}} &
\resultb{5}{9}{\MODELS{34}{2}}{\choices{295}{7}}{\conflicts{184}{5}}{\vars{494}{955}}{\constraints{1402}{2306}} &
\resultb{5}{10}{\MODELS{340}{2}}{\choices{660}{7}}{\conflicts{248}{4}}{\vars{589}{1097}}{\constraints{1717}{2715}} &
\resultb{5}{11}{\MODELS{2618}{2}}{\choices{3183}{8}}{\conflicts{437}{4}}{\vars{692}{1247}}{\constraints{2064}{3156}} &
\resultb{5}{12}{\MODELS{17204}{2}}{\choices{19209}{11}}{\conflicts{1791}{4}}{\vars{803}{1405}}{\constraints{2443}{3629}} &
choices \\ \cline{2-7}

&
\resultc{5}{8}{\MODELS{2}{2}}{\choices{141}{2}}{\conflicts{92}{1}}{\vars{407}{821}}{\constraints{1119}{1929}} &
\resultc{5}{9}{\MODELS{34}{2}}{\choices{295}{7}}{\conflicts{184}{5}}{\vars{494}{955}}{\constraints{1402}{2306}} &
\resultc{5}{10}{\MODELS{340}{2}}{\choices{660}{7}}{\conflicts{248}{4}}{\vars{589}{1097}}{\constraints{1717}{2715}} &
\resultc{5}{11}{\MODELS{2618}{2}}{\choices{3183}{8}}{\conflicts{437}{4}}{\vars{692}{1247}}{\constraints{2064}{3156}} &
\resultc{5}{12}{\MODELS{17204}{2}}{\choices{19209}{11}}{\conflicts{1791}{4}}{\vars{803}{1405}}{\constraints{2443}{3629}} &
constraints \\ \hline

\multirow{3}{*}{7} &
\resulta{7}{11}{\MODELS{2}{2}}{\choices{453}{2}}{\conflicts{327}{1}}{\vars{687}{1237}}{\constraints{2016}{3092}} &
\resulta{7}{12}{\MODELS{46}{2}}{\choices{842}{5}}{\conflicts{555}{4}}{\vars{798}{1395}}{\constraints{2391}{3561}} &
\resulta{7}{13}{\MODELS{598}{2}}{\choices{1758}{6}}{\conflicts{963}{5}}{\vars{917}{1561}}{\constraints{2798}{4062}} &
\resulta{7}{14}{\MODELS{5796}{2}}{\choices{7917}{7}}{\conflicts{1692}{6}}{\vars{1044}{1735}}{\constraints{3237}{4595}} &
\resulta{7}{15}{\MODELS{46690}{2}}{\choices{49982}{8}}{\conflicts{2571}{7}}{\vars{1179}{1917}}{\constraints{3708}{5160}} &
models \\ \cline{2-7}

&
\resultb{7}{11}{\MODELS{2}{2}}{\choices{453}{2}}{\conflicts{327}{1}}{\vars{687}{1237}}{\constraints{2016}{3092}} &
\resultb{7}{12}{\MODELS{46}{2}}{\choices{842}{5}}{\conflicts{555}{4}}{\vars{798}{1395}}{\constraints{2391}{3561}} &
\resultb{7}{13}{\MODELS{598}{2}}{\choices{1758}{6}}{\conflicts{963}{5}}{\vars{917}{1561}}{\constraints{2798}{4062}} &
\resultb{7}{14}{\MODELS{5796}{2}}{\choices{7917}{7}}{\conflicts{1692}{6}}{\vars{1044}{1735}}{\constraints{3237}{4595}} &
\resultb{7}{15}{\MODELS{46690}{2}}{\choices{49982}{8}}{\conflicts{2571}{7}}{\vars{1179}{1917}}{\constraints{3708}{5160}} &
choices \\ \cline{2-7}

&
\resultc{7}{11}{\MODELS{2}{2}}{\choices{453}{2}}{\conflicts{327}{1}}{\vars{687}{1237}}{\constraints{2016}{3092}} &
\resultc{7}{12}{\MODELS{46}{2}}{\choices{842}{5}}{\conflicts{555}{4}}{\vars{798}{1395}}{\constraints{2391}{3561}} &
\resultc{7}{13}{\MODELS{598}{2}}{\choices{1758}{6}}{\conflicts{963}{5}}{\vars{917}{1561}}{\constraints{2798}{4062}} &
\resultc{7}{14}{\MODELS{5796}{2}}{\choices{7917}{7}}{\conflicts{1692}{6}}{\vars{1044}{1735}}{\constraints{3237}{4595}} &
\resultc{7}{15}{\MODELS{46690}{2}}{\choices{49982}{8}}{\conflicts{2571}{7}}{\vars{1179}{1917}}{\constraints{3708}{5160}} &
constraints \\ \hline

\multirow{3}{*}{9} &
\resulta{9}{14}{\MODELS{2}{2}}{\choices{1560}{2}}{\conflicts{1206}{1}}{\vars{1039}{1725}}{\constraints{3181}{4523}} &
\resulta{9}{15}{\MODELS{58}{2}}{\choices{2206}{7}}{\conflicts{1669}{5}}{\vars{1174}{1907}}{\constraints{3648}{5084}} &
\resulta{9}{16}{\MODELS{928}{2}}{\choices{3437}{7}}{\conflicts{1934}{5}}{\vars{1317}{2097}}{\constraints{4147}{5677}} &
\resulta{9}{17}{\MODELS{10846}{2}}{\choices{15171}{7}}{\conflicts{3436}{5}}{\vars{1468}{2295}}{\constraints{4678}{6302}} &
\resulta{9}{18}{\MODELS{103530}{2}}{\choices{112964}{9}}{\conflicts{8331}{5}}{\vars{1627}{2501}}{\constraints{5241}{6959}} &
models \\ \cline{2-7}

&
\resultb{9}{14}{\MODELS{2}{2}}{\choices{1560}{2}}{\conflicts{1206}{1}}{\vars{1039}{1725}}{\constraints{3181}{4523}} &
\resultb{9}{15}{\MODELS{58}{2}}{\choices{2206}{7}}{\conflicts{1669}{5}}{\vars{1174}{1907}}{\constraints{3648}{5084}} &
\resultb{9}{16}{\MODELS{928}{2}}{\choices{3437}{7}}{\conflicts{1934}{5}}{\vars{1317}{2097}}{\constraints{4147}{5677}} &
\resultb{9}{17}{\MODELS{10846}{2}}{\choices{15171}{7}}{\conflicts{3436}{5}}{\vars{1468}{2295}}{\constraints{4678}{6302}} &
\resultb{9}{18}{\MODELS{103530}{2}}{\choices{112964}{9}}{\conflicts{8331}{5}}{\vars{1627}{2501}}{\constraints{5241}{6959}} &
choices \\ \cline{2-7}

&
\resultc{9}{14}{\MODELS{2}{2}}{\choices{1560}{2}}{\conflicts{1206}{1}}{\vars{1039}{1725}}{\constraints{3181}{4523}} &
\resultc{9}{15}{\MODELS{58}{2}}{\choices{2206}{7}}{\conflicts{1669}{5}}{\vars{1174}{1907}}{\constraints{3648}{5084}} &
\resultc{9}{16}{\MODELS{928}{2}}{\choices{3437}{7}}{\conflicts{1934}{5}}{\vars{1317}{2097}}{\constraints{4147}{5677}} &
\resultc{9}{17}{\MODELS{10846}{2}}{\choices{15171}{7}}{\conflicts{3436}{5}}{\vars{1468}{2295}}{\constraints{4678}{6302}} &
\resultc{9}{18}{\MODELS{103530}{2}}{\choices{112964}{9}}{\conflicts{8331}{5}}{\vars{1627}{2501}}{\constraints{5241}{6959}} &
constraints \\ \hline

\multirow{3}{*}{11} &
\resulta{11}{17}{\MODELS{2}{2}}{\choices{5057}{2}}{\conflicts{3982}{1}}{\vars{1463}{2285}}{\constraints{4614}{6222}} &
\resulta{11}{18}{\MODELS{70}{2}}{\choices{7896}{6}}{\conflicts{6259}{4}}{\vars{1622}{2491}}{\constraints{5173}{6875}} &
\resulta{11}{19}{\MODELS{1330}{2}}{\choices{7043}{7}}{\conflicts{4257}{4}}{\vars{1789}{2705}}{\constraints{5764}{7560}} &
\resulta{11}{20}{\MODELS{18200}{2}}{\choices{26276}{8}}{\conflicts{6374}{4}}{\vars{1964}{2927}}{\constraints{6387}{8277}} &
\resulta{11}{21}{\MODELS{200900}{2}}{\choices{219391}{9}}{\conflicts{15330}{4}}{\vars{2147}{3157}}{\constraints{7042}{9026}} &
models \\ \cline{2-7}

&
\resultb{11}{17}{\MODELS{2}{2}}{\choices{5057}{2}}{\conflicts{3982}{1}}{\vars{1463}{2285}}{\constraints{4614}{6222}} &
\resultb{11}{18}{\MODELS{70}{2}}{\choices{7896}{6}}{\conflicts{6259}{4}}{\vars{1622}{2491}}{\constraints{5173}{6875}} &
\resultb{11}{19}{\MODELS{1330}{2}}{\choices{7043}{7}}{\conflicts{4257}{4}}{\vars{1789}{2705}}{\constraints{5764}{7560}} &
\resultb{11}{20}{\MODELS{18200}{2}}{\choices{26276}{8}}{\conflicts{6374}{4}}{\vars{1964}{2927}}{\constraints{6387}{8277}} &
\resultb{11}{21}{\MODELS{200900}{2}}{\choices{219391}{9}}{\conflicts{15330}{4}}{\vars{2147}{3157}}{\constraints{7042}{9026}} &
choices \\ \cline{2-7}

&
\resultc{11}{17}{\MODELS{2}{2}}{\choices{5057}{2}}{\conflicts{3982}{1}}{\vars{1463}{2285}}{\constraints{4614}{6222}} &
\resultc{11}{18}{\MODELS{70}{2}}{\choices{7896}{6}}{\conflicts{6259}{4}}{\vars{1622}{2491}}{\constraints{5173}{6875}} &
\resultc{11}{19}{\MODELS{1330}{2}}{\choices{7043}{7}}{\conflicts{4257}{4}}{\vars{1789}{2705}}{\constraints{5764}{7560}} &
\resultc{11}{20}{\MODELS{18200}{2}}{\choices{26276}{8}}{\conflicts{6374}{4}}{\vars{1964}{2927}}{\constraints{6387}{8277}} &
\resultc{11}{21}{\MODELS{200900}{2}}{\choices{219391}{9}}{\conflicts{15330}{4}}{\vars{2147}{3157}}{\constraints{7042}{9026}} &
constraints \\ \hline

    \end{tabular}
\end{center}
\caption{Summary of experimental results in the elevator domain}
\label{tab:stats}
\end{table*}

%
Each entry, $a/c$, contrasts statistics obtained from the pure action theory in Line~1-14, viz.~$a$, with the
combination of action and control theory in Line~1-19, $c$.

For each setting,
we give the number of traces,
number of choices during search,
and the number of constraints in the solver (after all translations, preprocessing etc.).
First of all,
we notice that the addition of~\eqref{eq:elevator:control} yields exactly two valid traces,
no matter what setting is considered.
In the first trace, the elevator goes first \textit{up} all the way and then straight \textit{down},
and vice versa in the second trace.
Both traces are actually minimal as witnessed throughout the first column.
This is because the elevator is placed at the mid floor, otherwise one would be shorter than the other.
Next, we observe how drastically the number of trajectories and the underlying search increases for the action theories
with each extension of the horizon.
For instance, for 11 floors and an horizon of 19 (11+5+4) the mere action theory admits 200900 valid traces,
among which only two are conformant with~\eqref{eq:elevator:control}.
Looking at the underlying search, it is amazing how radically~\eqref{eq:elevator:control} trims the search,
in that only nine choices are needed to find the two traces.
This is also astonishing since its addition led to an increase from 7042 to~9026 constraints in the solver.
Similar yet less extreme observations can be made in all remaining settings.

To get an idea on runtime,
we conducted the same experiment on the larger instance with 71 floors ($\lambda$=107),
since the ones obtained for~5 to~11 floors are negligible.
Finding the first model without the dynamic constraint takes 19.4~sec, including 17.8~sec of solving,
while adding the dynamic constraint yields a runtime of only 2.2~sec with 0.01~sec of solving.
As with filtering traces, it seems that the dynamic constraint greatly contributes to guiding the solver.

Clearly, our empirical analysis is rather limited and can only indicate the potential impact of dynamic formulas
on reducing search efforts for finite traces.
Nonetheless,
we observe that adding the dynamic formula in~\eqref{eq:elevator:control}
not only (sometimes drastically) reduces the number of feasible traces
but also significantly cuts down the number of choices
despite its non-negligible increase of the resulting problem.


\section{Discussion}\label{sec:discussion}

We have elaborated upon the computational foundations of the Dynamic logic of Here-and-There and its equilibrium traces,
viz.\ \DHTf\ and \DELf~\cite{cadisc19a},
in order to design and implement an expressive ASP system for modeling and solving dynamic domains.
Our approach was motivated by the methodology of separating action and control theories, similar to what is done in Situation Calculus
and Golog~\cite{lereleli97a}.

To this end,
we carved out a normal form for dynamic formulas in \DELf\ that consists of its fragment corresponding to temporal logic programs. 
The translation of dynamic formulas into normal form heavily relies on the introduction of auxiliary variables.
This allows us to keep the size of the resulting temporal program polynomial in that of the original formula.
And moreover it allows us to overcome the common intranslatability of dynamic into temporal formulas when keeping the same language.
Our proof of the normal form result relies on a novel characterization of \DELf\ in terms of a three-valued logic.

The reduction of dynamic formulas to temporal logic programs enabled us to implement dynamic expression on top of the temporal ASP solver \telingo.
Since it constitutes a true extension of the ASP system \clingo,
we obtain a full-fledged modeling language extended by temporal and dynamic constructs.
We provided a limited empirical analysis demonstrating the potential impact of using dynamic formulas to select traces among the ones induced by an
associated temporal logic program.
This is how we see the interaction of control and action theories in our framework.

To the best of our knowledge,
\telingo\ provides the first ASP system augmented with constructs from dynamic and temporal logics.
Encodings for bounded model checking in \LTL\ over infinite traces were given in~\cite{helnie03a}.
This was extended to certain action theories expressed with dynamic operators in~\cite{gimadu13a}.
A key feature of these encodings is to capture loops inducing infinite traces.
This is avoided in~\cite{cabdie11a}, where infinite traces in \TEL\ are captured by B\"uchi automata via model checking.
Encodings of Golog in ASP were proposed in~\cite{sobanamc03a,ryan14a}.
This amounts to directly implementing a filter on traces, as done in Listing~\ref{lst:elevator},
without any logical underpinnings.
In~\cite{cadisc19a}, we provided a different translation from converse-free dynamic formulas in \DELf\ to propositional formulas in \HT,
which themselves can be translated into an equivalent disjunctive logic program (cf.\ \cite{capeva05a}).
More precisely, a dynamic formula $\gamma$ is translated in~\cite{cadisc19a} into a logic program $(\gamma)_i$.
This translation differs from the current one, $\sigma(\gamma)$, in several aspects.
First, the target language is different: while $\sigma(\gamma)$ produces a temporal logic program, $(\gamma)_i$ directly obtains a propositional logic program corresponding to some fixed time point $i$.
This is because $\sigma(\gamma)$ is thought for using \telingo\ as a backend, along with its incremental solving mode,
whereas $(\gamma)_i$ was thought for the direct use of a specific trace length.
%
Thus, the advantage of $(\gamma)_i$ is that it does not pass through temporal expressions from \telingo\ as an intermediate step.
On the other hand, its disadvantages are that $\gamma$ cannot contain the converse operator and that $(\gamma)_i$ can be exponential,
since it makes use of distributivity for normalization both of path expressions and of formulas into logic programs.
Note that $\sigma$ is applicable to any arbitrary dynamic formula $\gamma$ and takes polynomial time and space.
Its main disadvantage is that, in the general case, the resulting temporal logic program may not be amenable for incremental ASP computation.
To do so, an extra condition is required:
(non-constraint) temporal rules must additionally be \emph{present-centered}, that is,
conditions may refer to the past or present of head expressions, but not to their future.
In the general case, $\sigma(\gamma)$ may not satisfy this requirement:
for instance, one of the directions of \eqref{f:notcentered} yields the rule $\wnext \alwaysF (\previous \Lab{\alpha} \leftarrow \Lab{\beta})$,
so the head $\previous \Lab{\alpha}$ depends on a condition $\Lab{\beta}$ in its future.
Fortunately, in the case of negated formulas $\sigma(\neg\,\gamma)=\sigma(\bot \leftarrow \gamma)$, as the ones currently implemented in \telingo,
this limitation does not apply, since we can exclusively use constraints for the translation.
Our future work aims at a full integration of dynamic formulas into ASP and thus \telingo.
In particular, we will study the more general case in which dynamic expressions can be used in non-constraint rules.
As we did for temporal theories and the so-called \emph{past-future} form (see~\cite{cadisc19a}), we plan to identify a similar syntactic condition for a dynamic formula $\gamma$ so the resulting temporal program $\sigma(\gamma)$ is guaranteed to be present-centered.
It will be interesting to experiment with encodings deriving dynamic formulas rather than merely testing them.


\acknowledgements
This work was partially supported by Ministry of Science and Innovation, Spain (grant TIC2017-84453-P), Xunta de Galicia, Spain (grants GPC ED431B 2019/03 and 2016-2019 ED431G/01, CITIC Research Center).

 
\bibliographystyle{include/latex-class-ecai/ecai}
\newpage\appendix

\section{Proofs}
\label{sec:proofs}

\begin{proposition}
The valuations of derived formulas and path expressions correspond to:
\[
\begin{array}{rcl@{\hspace{10pt}}rcl}
&

&
\end{array}
\]
\begin{align*}
	\trival{k}{\varphi \wedge \psi} & \eqdef  \min\lbrace \trival{k}{\varphi},\trival{k}{\psi}\rbrace\\
	\trival{k}{\varphi \vee   \psi} & \eqdef  \max\rbrace\trival{k}{\varphi},\trival{k}{\psi}\rbrace\\
	\trival{k}{\varphi \to \psi} & \eqdef \begin{cases}
							2             & \text{if } \trival{k}{\varphi} \leq \trival{k}{\psi} \\
						\trival{k}{\psi}  & \text{otherwise}
	\end{cases}\\
	\trival{k}{\neg \varphi} & \eqdef 
	\begin{cases}
		2  & \text{if } \trival{k}{\varphi}=0 \\
		0  & \text{otherwise}
	\end{cases}\\
	\trival{k}{\initially} & \eqdef 
	\begin{cases}
		2 & \text{if } k=0 \\
		0 & \text{if } k>0
	\end{cases}	\\
    \trival{k}{\previous \varphi} & \eqdef 
	\begin{cases}
		0                      & \text{if } k=0 \\
		\trival{k-1}{\varphi}  & \text{if } k>0
	\end{cases}\\
	\trival{k}{\wprevious \varphi} & \eqdef 
	\begin{cases}
		2                      & \text{if } k=0 \\
		\trival{k-1}{\varphi}  & \text{if } k>0
	\end{cases}\\
\trival{k}{\varphi \since \psi} & \eqdef  \max\left\{\begin{array}{l} 
													 \min\lbrace \trival{j}{\psi},\trival{i}{\varphi}\mid \rangec{i}{j+1}{k}\rbrace\\ \mid\rangec{j}{0}{k} \end{array}\right\} \\
\trival{k}{\varphi \trigger \psi} & \eqdef 
	\min\left\{ \begin{array}{l}\max\lbrace \trival{j}{\psi},\trival{i}{\varphi}\mid \rangec{i}{j+1}{k}\rbrace\\ \mid \rangec{j}{0}{k} \end{array}\right\}\\			
	\trival{k}{\alwaysP \varphi} & \eqdef \min \lbrace\trival{i}{\varphi} \mid \rangec{i}{0}{k}\rbrace\\
	\trival{k}{\eventuallyP \varphi} & \eqdef \max\lbrace \trival{i}{\varphi} \mid \rangec{i}{0}{k}\rbrace\\
	\trival{k}{\finally} & \eqdef 
	\begin{cases}
		2 & \text{if } k+1=\lambda \\
		0 & \text{if } k+1<\lambda
	\end{cases}\\
	\trival{k}{\next \varphi} & \eqdef 
	\begin{cases}
		0                      & \text{if } k+1=\lambda \\
		\trival{k+1}{\varphi}  & \text{if } k+1<\lambda
	\end{cases}\\
	\trival{k}{\wnext \varphi} & \eqdef 
	\begin{cases}
		2                      & \text{if } k+1=\lambda \\
		\trival{k+1}{\varphi}  & \text{if } k+1<\lambda
	\end{cases}\\
\trival{k}{\varphi \until \psi} & \eqdef 
	\max\left\{\begin{array}{l}\min\lbrace \trival{j}{\psi},\trival{i}{\varphi}\mid \rangeo{i}{k}{j}\rbrace\\
	\mid \rangeo{j}{k}{\lambda} \end{array}\right\}\\
	\trival{k}{\varphi \release \psi} & \eqdef 
	\min\left\{\begin{array}{l}\max\lbrace \trival{j}{\psi},\trival{i}{\varphi}\mid \rangeo{i}{k}{j}\rbrace	\\ \mid
		 \rangeo{j}{k}{\lambda} \end{array}\right\}\\
	\trival{k}{\alwaysF \varphi} & \eqdef  \min \lbrace \trival{i}{\varphi} \mid \rangeo{i}{k}{\lambda} \rbrace \\
	\trival{k}{\eventuallyF \varphi} & \eqdef  \max\lbrace \trival{i}{\varphi} \mid \rangeo{i}{k}{\lambda}\rbrace\\
	\trival{k,j}{\rho^0} &\eqdef \begin{cases}
								2 & \text{if } j = k\\
								0 & \text{otherwise }
							 \end{cases}
	\\
    \trival{k,j}{\rho^{n+1}} &\eqdef
    \max \ \{ \min(\trival{k,i}{\rho},\trival{i,j}{\rho^n}) \mid \rangec{i}{k}{j} \}
\end{align*}

\end{proposition}

\begin{proofof}{Proposition~\ref{prop:three-valued}} By induction on the complexity of the formula. 
For the cases of $\top$ and $\bot$, note that

\begin{itemize}
\item $\tuple{\H,\T}, k \models \top$, $\tuple{\T,\T}, k \models \top$ and $\trival{k}{\top} = 2 \not = 0$.
\item $\tuple{\H,\T}, k \not \models \bot$, $\tuple{\T,\T}, k \not \models \bot$ and $\trival{k}{\bot} = 0 $.
\end{itemize}		

\noindent For the case of a propositional variable $p$, it is easy to check that 

\begin{itemize}
\item $\tuple{\H,\T}, k \models p$ iff $p \in \H_k$ iff $\trival{k}{p} = 2$;
\item $\tuple{\T,\T}, k \models p$ iff $p \in \T_k$ iff $\trival{k}{p} \not = 0$.
\end{itemize}

For the modal operators, we need to proceed by double induction. 
\begin{itemize}
\item \textbf{Case $\varphi=\DDia{\rho}\psi$: }
\begin{itemize}
\item \textbf{Item 1:} From left to right, if $\tuple{\H,\T}, k \models \DDia{\rho}\psi$, so $(k,j) \in \Rel{\rho}{\tuple{\H,\T}}$ and $\tuple{\H,\T}, j\models \psi$.
By induction we get $\trival{j}{\psi} = 2$ and $\trival{k,j}{\rho} = 2$.
Therefore, $\min\lbrace\trival{k,j}{\rho},\trival{j}{\psi}\rbrace= 2$.
Hence, $\max\lbrace \min\lbrace\trival{k,j}{\rho},\trival{j}{\psi}\rbrace  \mid 0 \le j < \lambda \rbrace = 2$.
By the satisfaction relation we get $\trival{k}{\DDia{\rho}\psi} = 2$.

Conversely, if $\trival{k}{\DDia{\rho}\psi} = 2$ then, by definition, there exists $0 \le j < \lambda$ such that $\min\lbrace \trival{k,j}{\rho}, \trival{j}{\psi}\rbrace = 2$.
Therefore, both $\trival{k,j}{\rho} = 2$ and $\trival{j}{\psi} = 2$.
By induction, $\tuple{\H,\T}, j \models \psi$ and $(k,j) \in \Rel{\rho}{\tuple{\H,\T}}$.
By the satisfaction relation, 
Therefore, $\tuple{\H,\T}, k \models \DDia{\rho}\psi$.

\item \textbf{Item 2:} From left to right, if $\tuple{\T,\T}, k \models \DDia{\rho}\psi$, then there exists $(k,j) \in \Rel{\rho}{\tuple{\T,\T}}$ and $\tuple{\T,\T}, j\models \psi$.
By induction we get $\trival{j}{\psi} \not= 0$ and  $\trival{k,j}{\rho} \not = 0$
Therefore, $\min\lbrace \trival{k,j}{\rho},\trival{j}{\psi}\rbrace \not = 0$.
Hence, $\max\lbrace \min\lbrace\trival{k,j}{\rho},\trival{j}{\psi}\rbrace  \mid 0 \le j < \lambda \rbrace \not = 0$.
By the (three-valued) satisfaction relation we get $\trival{k}{\DDia{\rho}\psi} \not = 0$.

From right to left, if $\trival{k}{\DDia{\rho}\psi} \not= 0$ then there exists $0 \le j < \lambda$ such that $\min\lbrace\trival{k,j}{\rho}, \trival{j}{\psi}\rbrace \not = 0$.
From this we conclude that $\trival{j}{\psi}\not = 0$ and  $\trival{k,j}{\rho} \not = 0$.
By induction $\tuple{\T,\T},j \models \psi$ and $(k,j) \in  \Rel{\rho}{\tuple{\T,\T}}$.
By the satisfaction relation we get $\tuple{\T,\T},k \models \DDia{\rho}\psi$.
\end{itemize}

\item \textbf{Case $\varphi=\DBox{\rho}\psi$: }
\begin{itemize}
\item \textbf{Item 1:} From left to right, assume by contradiction that $\trival{k}{\DBox{\rho}\psi} \not = 2$.
This means that $\min\lbrace \imp( \trival{k,j}{\rho} , \trival{j}{\psi}) \mid 0 \le j < \lambda \rbrace \not = 2$.
Therefore there exists $0\le j < \lambda$ such that $\imp( \trival{k,j}{\rho} , \trival{j}{\psi}) \not = 2$.
By definition, $ \trival{k,j}{\rho} > \trival{j}{\psi} \not = 2$.
We consider all cases:
\begin{itemize}
	\item If $\trival{j}{\psi} = 1 \not = 0$ then  $\trival{k,j}{\rho} = 2$. 
	By induction on $\psi$ and $\rho$ we get that $\tuple{\H,\T}, j \not \models \psi$ and $(k,j) \in  \Rel{\rho}{\tuple{\H,\T}}$.
	From the satisfaction relation we get  $\tuple{\H,\T}, k \not \models \DBox{\rho}\psi$: a contradiction.
	\item $\trival{j}{\psi} = 0$ then $\trival{k,j}{\rho} \in \lbrace 1,2\rbrace$ (so $\trival{k,j}{\rho} \not = 0$). By induction hypothesis $(k,j) \in \Rel{\rho}{\tuple{\T,\T}}$ and $\tuple{\T,\T}, j\not \models\psi$. Therefore, $\tuple{\T,\T}, k \not \models \DBox{\rho} \psi$.'
\end{itemize}

From right to left, let us assume by contradiction that $\tuple{\H,\T}, k \not \models \DBox{\rho}\psi$ and let us consider the following cases:

\begin{itemize}
\item There exists $(k,j) \in \Rel{\rho}{\tuple{\H,\T}}$ and $\tuple{\H,\T}, j \not \models \psi$.
By induction $\trival{k,j}{\rho}=2$ and $\trival{j}{\psi} \not = 2$.
Therefore, $\imp( \trival{k,j}{\rho} , \trival{j}{\psi}) = \trival{j}{\psi} \not = 2$.
By definition,  $\min\lbrace \imp( \trival{k,j}{\rho} , \trival{j}{\psi}) \mid 0 \le j < \lambda \rbrace \not = 2$.
Therefore $\trival{k}{\DBox{\rho}\psi} \not = 2$: a contradiction.

\item There exists $(k,j) \in \Rel{\rho}{\tuple{\T,\T}}$ and $\tuple{\T,\T}, j \not \models \psi$.
By induction $\trival{k,j}{\rho}\not = 0$ and $\trival{j}{\psi}=0$.
Therefore, $\imp( \trival{k,j}{\rho} , \trival{j}{\psi}) = \trival{j}{\psi} = 0$.
From this it follows that $\min\lbrace \imp( \trival{k,j}{\rho} , \trival{j}{\psi}) \mid 0 \le j < \lambda \rbrace  = 0$.
By the satisfaction relation we get $\trival{k}{\DBox{\rho}\psi} =0\not = 2$: a contradiction.
\end{itemize}
In any case we get a contradiction.

\item \textbf{Item 2:} From left to right, assume by contradiction that $\trival{k}{\DBox{\rho}\psi} = 0$.
Therefore $\min\lbrace \imp( \trival{k,j}{\rho} , \trival{j}{\psi}) \mid 0 \le j < \lambda \rbrace = 0$.
This means that there exists $0\le j < \lambda$ such that $\imp( \trival{k,j}{\rho} , \trival{j}{\psi}) = \trival{j}{\psi} = 0$.
Hence, $\trival{k,j}{\rho} \not = 0$ and $\trival{j}{\psi} = 0$.
By induction on $\rho$ and $\psi$ we conclude that $\tuple{\T,\T}, j \not \models \psi$ and $(k,j) \in \Rel{\rho}{\tuple{\T,\T}}$.
By the satisfaction relation it follows $\tuple{\T,\T},k\not \models \DBox{\rho} \psi$: a contradiction.

From right to left, assume by contradiction that $\tuple{\T,\T}, k \not \models \DBox{\rho}\psi$.
Therefore, there exists $(k,j) \in \Rel{\rho}{\tuple{\T,\T}}$ and $\tuple{\T,\T}, j \not \models \psi$.
By induction, $\trival{k,j}{\rho}\not = 0$ and $\trival{j}{\psi}=0$.
This means that $\imp(\trival{k,j}{\rho},\trival{j}{\psi}) = \trival{j}{\psi}=0$.
As a consequence, $\min\lbrace \imp( \trival{k,j}{\rho} , \trival{j}{\psi}) \mid 0 \le j < \lambda \rbrace = 0$, so $\trival{k}{\DBox{\rho}\psi} =0$: a contradiction.
\end{itemize}
\end{itemize}

\noindent In order to prove items $3$ and $4$ we proceed by induction on $\rho$.
\begin{itemize}
\item $\rho=\stp$:
\begin{itemize}
\item \textbf{Item $3$: } from left to right, if $(k,j)\in \Rel{\stp}{\tuple{\H,\T}}$ then $j = k+1$.
Therefore, $\trival{k,j}{\stp} = 2$.
Conversely, if $\trival{k,j}{\stp} = 2$ then $j=k+1$. By definition, $(k,j) \in \Rel{\stp}{\tuple{\H,\T}}$.

\item \textbf{Item $4$: } from left to right, if $(k,j)\in \Rel{\stp}{\tuple{\T,\T}}$ then $j = k+1$.
Therefore, $\trival{k,j}{\stp}= 2 \not = 0$.
Conversely, if $\trival{k,j}{\stp} \not = 0$ then $j=k+1$.
By definition $(k,j) \in \Rel{\stp}{\tuple{\T,\T}}$.
\end{itemize}

\item $\rho=\varphi?$:
\begin{itemize}
\item \textbf{Item $3$: } from left to right, if $(k,j)\in \Rel{\varphi?}{\tuple{\H,\T}}$ then $j = k$ and $\tuple{\H,\T}, k \models \varphi$.
By induction on $\varphi$ it follows that $\trival{k}{\varphi} = 2$ so $\trival{k,j}{\varphi?} = 2$.
Conversely, if $\trival{k,j}{\varphi? } = 2$ then $j=k$ and $\trival{k}{\varphi} = 2$.
By induction on $\varphi$ we get $\tuple{\H, \T},k \models \varphi$ so $(k,j) \in \Rel{\varphi?}{\tuple{\H,\T}}$.
\item \textbf{Item $4$: } from left to right, if $(k,j)\in \Rel{\varphi?}{\tuple{\T,\T}}$ then $j = k$ and $\tuple{\T,\T}, k \models \varphi$.
By induction on $\varphi$,  $\trival{k}{\varphi} \not = 0$ so $\trival{k,j}{\varphi?} \not = 0$.
Conversely, if $\trival{k,j}{\varphi?} \not = 0$ then $j=k$ and $\trival{k}{\varphi} \not = 0$.
By induction on $\varphi$ we get $\tuple{\T, \T},k \models \varphi$ so $(k,j) \in \Rel{\varphi?}{\tuple{\T,\T}}$.
\end{itemize}

\item $\rho=\rho_1 + \rho_2$:
\begin{itemize}
\item \textbf{Item $3$: } from left to right, if $(k,j) \in \Rel{\rho_1 + \rho_2}{\tuple{\H,\T}}$ then either $(k,j) \in \Rel{\rho_1}{\tuple{\H,\T}}$ or $(k,j) \in \Rel{\rho_2}{\tuple{\H,\T}}$. By induction on $\rho_1$ and $\rho_2$ we get that either $\trival{k,j}{\rho_1} = 2$ or $\trival{k,j}{\rho_2} = 2$.
Therefore $\max\lbrace\trival{k,j}{\rho_1},\trival{k,j}{\rho_2}\rbrace = 2$, so $\trival{k,j}{\rho_1+\rho_2} = 2$.
Conversely, if $\trival{k,j}{\rho_1+\rho_2} = 2$ then $\max\lbrace\trival{k,j}{\rho_1},\trival{k,j}{\rho_2}\rbrace = 2$, so either  $\trival{k,j}{\rho_1} = 2$ or $\trival{k,j}{\rho_2} = 2$.
By induction hypothesis we get that either  $(k,j) \in \Rel{\rho_1}{\tuple{\H,\T}}$ or $(k,j) \in \Rel{\rho_2}{\tuple{\H,\T}}$, so  $(k,j) \in \Rel{\rho_1+\rho_2}{\tuple{\H,\T}}$.

\item \textbf{Item $4$: } from left to right, if $(k,j) \in \Rel{\rho_1 + \rho_2}{\tuple{\T,\T}}$ then either $(k,j) \in \Rel{\rho_1}{\tuple{\T,\T}}$ or $(k,j) \in \Rel{\rho_2}{\tuple{\T,\T}}$. By induction on $\rho_1$ and $\rho_2$ we get that either $\trival{k,j}{\rho_1} \not= 0$ or $\trival{k,j}{\rho_2} \not = 0$.
Therefore $\max\lbrace \trival{k,j}{\rho_1},\trival{k,j}{\rho_2}\rbrace \not = 0$, so $\trival{k,j}{\rho_1+\rho_2} \not = 0$.
Conversely, if $\trival{k,j}{\rho_1+\rho_2} \not = 0$ then $\max\lbrace\trival{k,j}{\rho_1},\trival{k,j}{\rho_2}\rbrace \not =  0$, so either  $\trival{k,j}{\rho_1} \not =  0$ or $\trival{k,j}{\rho_2} \not =  0$.
By induction on $\rho_1$ and $\rho_2$ we get that either  $(k,j) \in \Rel{\rho_1}{\tuple{\T,\T}}$ or $(k,j) \in \Rel{\rho_2}{\tuple{\T,\T}}$, so  $(k,j) \in \Rel{\rho_1+\rho_2}{\tuple{\T,\T}}$.
\end{itemize}

\item $\rho = \rho_1;\rho_2$:
\begin{itemize}
\item \textbf{Item $3$: } from left to right, if $(k,j)\in \Rel{\rho_1;\rho_2}{\tuple{\H,\T}}$ then there exists $\rangeo{i}{0}{\lambda}$ such that $(k,i)\in \Rel{\rho_1}{\tuple{\H,\T}}$ and $(i,j)\in \Rel{\rho_2}{\tuple{\H,\T}}$.
By induction we get that $\trival{k,i}{\rho_1} = 2$ and $\trival{i,j}{\rho_2} = 2$.
Therefore $\min\lbrace\trival{k,i}{\rho_1}, \trival{i,j}{\rho_2}\rbrace = 2$.
By definition $\trival{k,j}{\rho_1;\rho_2} = 2$.
Conversely, if $\trival{k,j}{\rho_1;\rho_2} = 2$ then there exists $\rangeo{i}{0}{\lambda}$ such that $\min\lbrace\trival{k,i}{\rho_1},\trival{i,j}{\rho_2}\rbrace = 2$.
This means that $\trival{k,i}{\rho_1} = 2$ and $\trival{i,j}{\rho_2} = 2$.
By induction hypothesis $(k,i)\in \Rel{\rho_1}{\tuple{\H,\T}}$ and $(i,j)\in \Rel{\rho_2}{\tuple{\H,\T}}$.
By definition $(k,j)\in \Rel{\rho_1;\rho_2}{\tuple{\H,\T}}$.
\item \textbf{Item $4$: } from left to right, if $(k,j)\in \Rel{\rho_1;\rho_2}{\tuple{\T,\T}}$ then there exists $\rangeo{i}{0}{\lambda}$ such that $(k,i)\in \Rel{\rho_1}{\tuple{\T,\T}}$ and $(i,j)\in \Rel{\rho_2}{\tuple{\T,\T}}$.
By induction we get that $\trival{k,i}{\rho_1} \not= 0$ and $\trival{i,j}{\rho_2} \not = 0$.
Therefore $\min\lbrace\trival{k,i}{\rho_1}, \trival{i,j}{\rho_2}\rbrace \not = 0$.
By definition $\trival{k,j}{\rho_1;\rho_2} \not = 0$.
Conversely, if $\trival{k,j}{\rho_1;\rho_2} \not = 0$ then there exists $\rangeo{i}{0}{\lambda}$ such that $\min\lbrace\trival{k,i}{\rho_1},\trival{i,j}{\rho_2}\rbrace \not = 0$.
This means that $\trival{k,i}{\rho_1} \not = 0$ and $\trival{i,j}{\rho_2} \not = 0$.
By induction hypothesis $(k,i)\in \Rel{\rho_1}{\tuple{\T,\T}}$ and $(i,j)\in \Rel{\rho_2}{\tuple{\T,\T}}$.
By definition $(k,j)\in \Rel{\rho_1;\rho_2}{\tuple{\T,\T}}$.
\end{itemize}

\item $\rho = \rho^n$:
We proceed by induction on $n$ in its turn.
For $n=0$ we have that $(k,j) \in \Rel{\rho^0}{\tuple{\H,\T}} \ \hbox{ iff } \ k=j \ \hbox{ iff }\ \trival{k,j}{\rho^0}=2$. 
Suppose proved it up to $n\geq 0$.
Then, $(k,j) \in \Rel{\rho^{n+1}}{\tuple{\H,\T}}$ is equivalent to:
\[
\exists i \ \text{s.t. } \rangeo{i}{0}{\lambda}: (k,i) \in \Rel{\rho}{\tuple{\H,\T}} \ \text{and } (i,j) \in \Rel{\rho^n}{\tuple{\H,\T}}
\]
by structural induction for $\rho$ and induction on $n$ this is equivalent to:
\[
\begin{array}{rcl}
& & \exists i \ \text{s.t. } \rangeo{i}{0}{\lambda}: \trival{k,i}{\rho}=2 \ \text{and } \trival{i,j}{\rho^n}=2\\
& \hbox{ iff } &
\exists i \ \text{s.t. } \rangeo{i}{0}{\lambda}: \min(\trival{k,i}{\rho},\trival{i,j}{\rho^n})=2\\
& \hbox{ iff } &
\max\lbrace \min\lbrace\trival{k,i}{\rho},\trival{i,j}{\rho^n}\rbrace \mid i=k..j\rbrace = 2\\
& \hbox{ iff } &
\trival{k,i}{\rho^{n+1}}= 2
\end{array}
\]
The case for $(k,j)\in \Rel{\rho^n}{\tuple{\T,\T}}$ iff $\trival{k,j}{\rho^{n}}\neq 0$ is analogous.

\item $\rho = \rho^*$:
\[
\begin{array}{rcll}
& & (k,j)\in \Rel{\rho^*}{\tuple{\H,\T}}\\
& \hbox{ iff } &
\exists n\geq 0 \ \text{s.t. } (k,j)\in \Rel{\rho^n}{\tuple{\H,\T}}\\
& \hbox{ iff } &\exists n\geq 0 \ \text{s.t. } \trival{k,j}{\rho^{n}}= 2 & \text{(IH)}\\
& \hbox{ iff } &\max\lbrace\trival{k,j}{\rho^{n}} \mid n\geq0\rbrace= 2 \\
& \hbox{ iff }& \trival{k,j}{\rho^*}= 2 
\end{array}
\]
The case for $(k,j)\in \Rel{\rho^*}{\tuple{\T,\T}}$ iff $\trival{k,j}{\rho^{*}}\neq 0$ is analogous.

\end{itemize}
\end{proofof}

\begin{proposition}\label{prop:equivalences}
	The following expressions are \DHTf{}-valid
	\begin{eqnarray}
	\DBox{\rho_1 + \rho_2}\varphi &\leftrightarrow& \DBox{\rho_1} \varphi \wedge \DBox{\rho_2} \varphi \label{eq:valid:1}\\
	\DDia{\rho_1 + \rho_2}\varphi &\leftrightarrow& \DDia{\rho_1} \varphi \vee \DDia{\rho_2} \varphi \label{eq:valid:2}\\
	\DBox{\rho_1;\rho_2}\varphi &\leftrightarrow& \DBox{\rho_1}\DBox{\rho_2} \varphi \label{eq:valid:3}\\
	\DDia{\rho_1;\rho_2}\varphi &\leftrightarrow& \DDia{\rho_1}\DDia{\rho_2} \varphi \label{eq:valid:4}\\
	\DBox{\rho^*} \varphi &\leftrightarrow& \varphi \wedge \DBox{\rho}\DBox{\rho^*}\varphi \label{eq:valid:5}\\
	\finally &\rightarrow& \left( \DBox{\rho^*} \varphi \leftrightarrow \varphi \right)\label{eq:valid:6}\\
	\DDia{\rho^*} \varphi &\leftrightarrow& \varphi \vee \DDia{\rho}\DDia{\rho^*}\varphi\label{eq:valid:7}\\
	\finally &\rightarrow& \left( \DDia{\rho^*} \varphi \leftrightarrow \varphi \right)\label{eq:valid:8}\\
	\DDia{\psi?} \varphi &\leftrightarrow& \psi \wedge \varphi\label{eq:valid:9}\\
	\DBox{\psi?} \varphi &\leftrightarrow& \left(\psi \rightarrow \varphi\right)\label{eq:valid:10}\\
	\wnext\varphi &\leftrightarrow& \DBox{\stp} \varphi\label{eq:valid:11}\\
	\next\varphi &\leftrightarrow& \DDia{\stp} \varphi\label{eq:valid:13}\\
	\wprevious\varphi &\leftrightarrow& \DBox{\stp^-} \varphi \label{eq:valid:16}\\
	\previous\varphi &\leftrightarrow& \DDia{\stp^-} \varphi\label{eq:valid:17}		
	\end{eqnarray}
\end{proposition}

\begin{proofof}{Lemma~\ref{lem:nf1}} 		
Take the \DHTf{}-trace $\tuple{\H',\T'}$ whose three valued interpretation ${\bm m}'$ satisfies:
\begin{align*}
\trivalp{k}{\Lab{\varphi}} \overset{\label{proofNF_equalityExtAlph}}{=} \trival{k}{\varphi}
\end{align*}
for any formula $\varphi$ over $\mathcal{A}$ and for all $\rangeo{i}{k}{\lambda}$.
When $\varphi$ is an atom $a \in \mathcal{A}$ then $\trivalp{k}{a} = \trivalp{k}{\Lab{a}} = \trival{k}{a}$, which implies that both valuations coincide for atoms, and so, $\tuple{\H',\T'} |_{\mathcal{A}} = \tuple{\H,\T}$.
It remains to be shown that
$\tuple{\H',\T'} \models \sigma(\Gamma)$,
which is equivalent to
\begin{align*}
\tuple{\H',\T'} &\models
\left\lbrace \Lab{\gamma} \mid \gamma \in \Gamma \right\rbrace
\cup \left\lbrace \eta(\mu)
\mid \mu \in \mathit{FL}(\Gamma)\right\rbrace\\
\Leftrightarrow \tuple{\H',\T'} &\models
\left\lbrace \Lab{\gamma} \mid \gamma \in \Gamma \right\rbrace
\text{ and }
\tuple{\H',\T'} \models \left\lbrace \eta(\mu)
\mid \mu \in \mathit{FL}(\Gamma)\right\rbrace
\end{align*}

The first satisfaction relation follows directly from the definition of $\tuple{\H',\T'}$ since $\trivalp{0}{\Lab{\gamma}}=2$ iff $\trival{0}{\gamma}=2$ and we had that $\tuple{\H,\T}$ is a model of $\Gamma$.
For the second part, we consider the following cases depending on the structure of the subformula $\mu$:

\begin{enumerate}
	\item For $\mu = \DDia{\stp} \varphi$ we have two formulas in $\eta(\mu)$
	
	\begin{itemize}
		\item For the formula $\wnext\alwaysF (\previous\Lab{\mu} \leftrightarrow \Lab{\varphi})$, the equivalence must be satisfied for any $\rangeo{k}{1}{\lambda}$ ( $\lambda=1$ being trivial).
		Then,
		\begin{align*}
		\trivalp{k}{\previous\Lab{\mu}} &= \trivalp{k-1}{\Lab\mu} = \trival{k-1}{\mu}\\
										&= \trival{k-1}{\DDia{\stp}\varphi} = \trival{k}{\varphi} = \trivalp{k}{\Lab\varphi}
		\end{align*}
		\item For the second formula, $\alwaysF (\finally \to \neg \Lab{\mu})$ we get the following 
		                
		\begin{align*}
		2&=\trivalp{0}{\alwaysF (\finally \to \neg \Lab{\mu})} = \trivalp{\lambda-1}{\finally \to \neg \Lab{\mu}}\\
															& =  \trivalp{\lambda-1}{\neg \Lab{\mu}}.
		\end{align*} 
		
		\noindent Note on the other side that, $0 = \trival{\lambda -1}{\DDia{\stp}\varphi} = \trival{\lambda -1}{\mu} = \trivalp{\lambda -1}{\Lab\mu}$. Therefore,  $\trivalp{\lambda-1}{\neg \Lab{\mu}} = 2$.
	\end{itemize}
	
	\item For $\mu = \DBox{\stp} \varphi$ we have two formulas in $\eta(\mu)$. For the formula $\wnext\alwaysF (\previous\Lab{\mu} \leftrightarrow \Lab{\varphi})$ we refer the reader to the case of $\DDia{\stp}\varphi$. For the second formula, $\alwaysF (\finally \to \Lab{\mu})$, we present the proof below:

				\begin{align*}
				\trivalp{0}{\alwaysF (\finally \to \Lab{\mu})}  &= \trivalp{\lambda-1}{\finally \to \Lab{\mu}} =  \trivalp{\lambda-1}{\Lab{\mu}} \\
				&=\trival{\lambda-1}{\mu}= \trival{\lambda-1}{\DBox{\stp} \varphi}.
				\end{align*} 
	
	\item For $\mu = \DDia{\stp^-} \varphi$: we have two formulas in $\eta(\mu)$:
	
	\begin{itemize}
		\item[-] For the formula $\wnext \alwaysF(\Lab\mu \leftrightarrow \previous \Lab\varphi)$ note that the prefix $\wnext \alwaysF$ means that the double implication must be satisfied for any $\rangeo{k}{1}{\lambda}$ and, moreover, that this is trivially true when $\lambda = 1$.
		So, we have to prove $\trivalp{k}{\Lab\mu}=\trivalp{k}{\previous \Lab\varphi}$ for all $k=1..n$ and may assume $n>0$.
		The proof can be obtained as follows:
		
		\begin{align*}
		\trivalp{k}{\Lab{\mu}}& = \trival{k}{\mu} = \trival{k}{\DDia{\stp^-} \varphi} \stackrel{\eqref{eq:valid:17}}{=} \trival{k}{\previous \varphi}\\
							  & = \trival{k-1}{\varphi} =  \trival{k-1}{\Lab\varphi} =  \trival{k}{\previous \Lab \varphi}.
		\end{align*}
		\item[-] For satisfying the formula $\neg \Lab\mu$, this is the same than requiring $\trivalp{0}{\Lab\mu}=0$ and this follows from
		
		\begin{displaymath}
		\trivalp{0}{\Lab\mu}=\trival{0}{\mu}=\trival{0}{\DDia{\stp^-} \varphi}=0
		\end{displaymath}
	\end{itemize}
	
	\item For $\mu = \DBox{\stp^-} \varphi$ we have two formulas in $\eta(\mu)$. For the formula $\wnext \alwaysF(\Lab\mu \leftrightarrow \previous \Lab\varphi)$ we refer the reader to the case of $\DDia{\stp^-}\varphi$. For the second formula, $\Lab \mu$, note that $\trivalp{0}{\Lab\mu} = \trival{0}{\mu} = 2$ by definition.
	
	\item For $\mu = \DDia{\psi?} \varphi$ we have $\eta(\mu) = \alwaysF (\Lab{\mu} \leftrightarrow \Lab{\varphi} \wedge \Lab{\psi})$ and so, $\tuple{\H',\T'} \models \eta(\mu)$ amounts to proving $\trivalp{k}{\Lab{\mu}}=\trivalp{k}{\Lab{\varphi} \wedge \Lab{\psi}}$ for all $\rangeo{k}{0}{\lambda}$. In this case we have that
	
	\begin{align*}
	\trivalp{k}{\Lab{\mu}} = \trival{k}{\mu} &= \trival{k}{ \DDia{\psi?}\varphi} \stackrel{\eqref{eq:valid:9}}{=} \trival{k}{\psi \wedge \varphi}\\
	 &= \min\lbrace\trival{k}{\psi},\trival{k}{\varphi}\rbrace\\
	&= \min\lbrace\trivalp{k}{\Lab{\psi}},\trivalp{k}{\Lab{\varphi}}\rbrace\\ 
	&= \trivalp{k}{\Lab{\psi} \wedge \Lab{\varphi}}.\\
	\end{align*}
	
	\item For $\mu = \DBox{\psi?} \varphi$ we have that $\eta(\mu) = \alwaysF (\Lab{\mu} \leftrightarrow (\Lab{\psi} \rightarrow \Lab{\varphi}))$ and so, $\tuple{\H',\T'} \models \eta(\mu)$ amounts to proving $\trivalp{k}{\Lab{\mu}}=\trivalp{k}{\Lab{\psi} \rightarrow \Lab{\varphi}}$ for all $\rangeo{k}{0}{\lambda}$. 
	In this case we have
	
	\begin{align*}
	\trivalp{k}{\Lab{\mu}} &= \trival{k}{\mu}= \trival{k}{\DBox{\psi?}\varphi} \stackrel{\eqref{eq:valid:10}}{=} \trival{k}{\psi \rightarrow \varphi}\\
	&= \begin{cases}
	2 & \hbox{ if } \trival{k}{\psi} \le \trival{k}{\varphi}\\
	\trival{k}{\varphi} & \text{otherwise}
	\end{cases}\\
	&= \begin{cases}
	2 & \hbox{ if } \trivalp{k}{\Lab{\psi}} \le \trivalp{k}{\Lab{\varphi}}\\
	\trivalp{k}{\Lab{\varphi}} & \text{otherwise}
	\end{cases}\\
	& = \trivalp{k}{\Lab{\psi} \rightarrow \Lab{\varphi}}.
	\end{align*}

	\item For $\mu = \DDia{\rho_1 + \rho_2} \varphi$  we have $\eta(\mu) = \alwaysF (\Lab{\mu} \leftrightarrow \Lab\alpha \vee \Lab\beta)$ (with $\alpha = \DDia{\rho_1}\varphi$ and $\beta = \DDia{\rho_2}\varphi$) and so, $\tuple{\H',\T'} \models \eta(\mu)$ amounts to proving $\trivalp{k}{\Lab{\mu}}=\trivalp{k}{\Lab\alpha \vee \Lab\beta}$ for all $\rangeo{k}{0}{\lambda}$. 
	In this case we have
	
	\begin{align*}
	\trivalp{k}{\Lab{\mu}} & = \trival{k}{\mu} = \trival{k}{ \DDia{\rho_1 + \rho_2}\varphi}\\
						   & \stackrel{\eqref{eq:valid:2}}{=} \trival{k}{\DDia{\rho_1}\varphi \vee \DDia{\rho_2}\varphi}\\
						   &= \max\lbrace\trival{k}{ \DDia{\rho_1}\varphi},\trival{k}{\DDia{\rho_2}\varphi}\rbrace \\
						   &= \max\lbrace\trival{k}{\alpha},\trival{k}{\beta}\rbrace \\
						   &\stackrel{\alpha,\beta \in FL(\mu)}{=} \max\lbrace\trivalp{k}{\Lab\alpha},\trivalp{k}{\Lab\beta}\rbrace\\
						   &= \trivalp{k}{\Lab\alpha \vee \Lab\beta}
	\end{align*}

	\item For $\mu = \DBox{\rho_1 + \rho_2} \varphi$  we have $\eta(\mu) = \alwaysF (\Lab{\mu} \leftrightarrow \Lab\alpha \wedge \Lab\beta)$ (with $\alpha = \DBox{\rho_1}\varphi$ and $\beta = \DBox{\rho_2}\varphi$) and so, $\tuple{\H',\T'} \models \eta(\mu)$ amounts to proving $\trivalp{k}{\Lab{\mu}}=\trivalp{k}{\Lab\alpha \wedge \Lab\beta}$ for all $\rangeo{k}{0}{\lambda}$. 
	In this case we have
	
	\begin{align*}
	\trivalp{k}{\Lab{\mu}} & = \trival{k}{\mu} = \trival{k}{ \DBox{\rho_1 + \rho_2}\varphi}\\
	& \stackrel{\eqref{eq:valid:1}}{=} \trival{k}{\DBox{\rho_1}\varphi \wedge \DBox{\rho_2}\varphi}\\
	&= \min\lbrace\trival{k}{ \DBox{\rho_1}\varphi},\trival{k}{\DBox{\rho_2}\varphi}\rbrace \\
	&= \min\lbrace\trival{k}{\alpha},\trival{k}{\beta}\rbrace \\
	&\stackrel{\alpha,\beta \in FL(\mu)}{=} \min\lbrace\trivalp{k}{\Lab\alpha},\trivalp{k}{\Lab\beta}\rbrace\\
	&= \trivalp{k}{\Lab\alpha \wedge \Lab\beta}
	\end{align*}

\item For $\mu = \DDia{\rho_1;\rho_2} \varphi$, $\tuple{\H',\T'} \models\eta(\mu)$ amounts to proving $\trivalp{k}{\Lab{\mu}}=\trivalp{k}{\Lab \alpha}$, for $\rangeo{k}{0}{\lambda}$, with $\alpha = \DDia{\rho_1}\DDia{\rho_2}\varphi$:

\begin{align*}
\trivalp{k}{\Lab{\mu}} &= \trival{k}{\mu} = \trival{k}{\DDia{\rho_1;\rho_2}\varphi} \stackrel{\eqref{eq:valid:4}}{=} \trival{k}{\DDia{\rho_1}\DDia{\rho_2}\varphi}\\		
					   & = \trival{k}{\alpha} \stackrel{\alpha \in FL(\mu)}{=} \trivalp{k}{\Lab{\alpha}}
\end{align*}

\item For $\mu = \DBox{\rho_1;\rho_2} \varphi$, $\tuple{\H',\T'} \models\eta(\mu)$ amounts to proving $\trivalp{k}{\Lab{\mu}}=\trivalp{k}{\Lab \alpha}$, for $\rangeo{k}{0}{\lambda}$, with $\alpha = \DBox{\rho_1}\DBox{\rho_2}\varphi$:

\begin{align*}
\trivalp{k}{\Lab{\mu}} &= \trival{k}{\mu} = \trival{k}{\DBox{\rho_1;\rho_2}\varphi} \stackrel{\eqref{eq:valid:3}}{=} \trival{k}{\DBox{\rho_1}\DBox{\rho_2}\varphi}\\		
& = \trival{k}{\alpha} \stackrel{\alpha \in FL(\mu)}{=} \trivalp{k}{\Lab{\alpha}}
\end{align*}

	\item For $\mu = \DDia{\rho^*} \varphi$ we have two formulas in $\eta(\mu)$.
	
	\begin{itemize}
		\item For the formula $\alwaysF( \Lab{\mu} \leftrightarrow \Lab{\varphi} \vee \Lab{\alpha}$, where $\alpha = \DDia{\rho}\DDia{\rho*}\mu$, it amounts to show that $\trivalp{k}{ \Lab{\mu}} = \trivalp{k}{ \Lab{\varphi} \vee  \Lab{\alpha}}$ for all $\rangeo{k}{0}{\lambda}$:
		
		\begin{align*}
		&\trivalp{k}{ \Lab{\mu}} = \trival{k}{\mu}= \trival{k}{\DDia{\rho^*} \varphi} \stackrel{\eqref{eq:valid:7}}{=} \trival{k}{\varphi \vee \DDia{\rho}\DDia{\rho^*} \varphi}\\
		&=  \max\lbrace\trival{k}{\varphi},\trival{k}{\alpha}\rbrace\\
		&\stackrel{\alpha \in FL(\mu)}{=} \max\lbrace\trivalp{k}{\Lab\varphi},\trivalp{k}{\Lab\alpha}\rbrace\\
		&= \trivalp{k}{\Lab\varphi \vee \Lab\alpha}.						   							   
		\end{align*}
								
		\item For the formula $\alwaysF(\finally \to ( \Lab{\mu} \leftrightarrow \Lab{\varphi}))$ the proof amounts to prove that $\trivalp{\lambda-1}{\Lab\mu} = \trivalp{\lambda-1}{\Lab\varphi}$.
		We show this below.
		\begin{align*}
		&\trivalp{\lambda-1}{\Lab\mu} = \trival{\lambda-1}{\mu} = \trival{\lambda-1}{\DDia{\rho^*} \varphi}\\
		&\stackrel{\eqref{eq:valid:8}}{=}  \trival{\lambda-1}{\varphi}= \trivalp{\lambda-1}{\Lab \varphi}.						
		\end{align*}
	\end{itemize}

	\item For $\mu = \DBox{\rho^*} \varphi$ we have two formulas in $\eta(\mu)$.
	
	\begin{itemize}
		\item For the formula $\alwaysF( \Lab{\mu} \leftrightarrow \Lab{\alpha} \wedge \Lab{\beta}$, where $\alpha = \DBox{\rho}\DBox{\rho*}\mu$, it amounts to show that $\trivalp{k}{ \Lab{\mu}} = \trivalp{k}{ \Lab{\varphi} \wedge  \Lab{\alpha}}$ for all $\rangeo{k}{0}{\lambda}$:
		
		\begin{align*}
		&\trivalp{k}{ \Lab{\mu}} = \trival{k}{\mu}= \trival{k}{\DBox{\rho^*} \varphi} \stackrel{\eqref{eq:valid:6}}{=} \trival{k}{\varphi \wedge \DBox{\rho}\DBox{\rho^*} \varphi}\\
		&=  \min\lbrace\trival{k}{\varphi},\trival{k}{\alpha}\rbrace\\
		&\stackrel{\alpha \in FL(\mu)}{=} \min\lbrace\trivalp{k}{\Lab\varphi},\trivalp{k}{\Lab\alpha}\rbrace\\
		&= \trivalp{k}{\Lab\varphi \wedge \Lab\alpha}.						   							   
		\end{align*}
		
		\item For the formula $\alwaysF(\finally \to ( \Lab{\mu} \leftrightarrow \Lab{\varphi}))$, we refer the reader to the previous case (but using ~\eqref{eq:valid:6} instead).		
	\end{itemize}	
\end{enumerate}
\end{proofof}

\begin{proofof}{Lemma~\ref{lem:nf2}}
We proceed by structural induction on $\mu$.
\begin{enumerate}
	\item If $\mu$ is a propositional variable $p$, $\bot$ or $\top$, the proof is trivial because $\Lab\mu=\mu$ by definition.
	\item If $\mu = \DDia{\stp} \varphi$, we divide the proof in two cases:
	\begin{itemize}
		\item[-] If $k=\lambda-1$ we use the second formula in $\eta(\mu)$.
		It follows that
		\begin{align*}
		2 &= \trival{\lambda-1}{\finally \rightarrow \neg \Lab{\mu}} =  \trival{\lambda-1}{\neg \Lab{\mu}}\\
		 & \hbox{ iff } \trival{\lambda-1}{\Lab{\mu}} = 0 = \trival{\lambda-1}{\DDia{\stp} \varphi}.
		\end{align*}
			
		\item[-] If $0\leq k < \lambda-1$ we can apply the first formula in $\eta(\mu)$ that guarantees $\trival{j}{\previous \Lab\mu}=\trival{j}{\Lab\varphi}$ for all $j=1..\lambda-1$.
		In particular, we can take $j=k$ and so:
		
		\begin{align*}
		\trival{k}{\Lab\mu} & = \trival{k+1}{\previous \Lab\mu} \stackrel{\eta(\mu)}{=} \trival{k+1}{\Lab\varphi}\\
							& \stackrel{\text{IH}}{=} \trival{k+1}{\varphi} = \trival{k}{\next \varphi} \stackrel{\eqref{eq:valid:13}}{=} \trival{k}{\DDia{\top} \varphi}.
		\end{align*}
	\end{itemize}

	\item If $\mu = \DBox{\stp} \varphi$, we consider two cases
	
	\begin{itemize}
		\item $k = \lambda-1$: on one side we take the second formula of $\eta(\mu)$ to conclude that $\trival{\lambda -1}{\Lab \mu} = 2$ and, by definition, $\trival{\lambda -1}{\mu} = 2$.
					
		\item if $0 \le k < \lambda-1$  we refer the reader to the previous case.		
	\end{itemize}

	\item If $\mu = \DDia{\stp^-} \varphi$ we divide into two cases:
	
	\begin{itemize}
		\item[-] If $k=0$ we directly use the second formula in $\eta(\mu)$ to conclude
		
		\begin{displaymath}
		2 = \trival{0}{\neg \Lab{\mu}} \hbox{ iff } \trival{0}{\Lab{\mu}} = 0 = \trival{0}{\previous \varphi}.
		\end{displaymath}
		
		\item[-] If $k>0$ we can apply the first formula in $\eta(\mu)$ as follows:
		\begin{align*}
		\trival{k}{\Lab\mu} &= \trival{k}{\previous \Lab\varphi} \stackrel{k>0}{=} \trival{k-1}{\Lab\varphi} \\
							& \stackrel{\text{IH}}{=} \trival{k-1}{\varphi} = \trival{k}{\previous \varphi}\\ 
							& \stackrel{\eqref{eq:valid:17}}{=} \trival{k}{\DDia{\stp^-}\varphi}\\
		\end{align*}
	\end{itemize}
	
	\item $\mu = \DBox{\stp^-} \varphi$ we distinguish two cases
	
	\begin{itemize}
		\item[-] $k = 0$ we use the second formula in $\eta(\mu)$ to conclude that $2= \trivalp{0}{\Lab \mu} = \trival{0}{\mu}$
		\item[-] $k > 0$ we use the first formula in $\eta(\mu)$ as follows

	\begin{align*}
		\trival{k}{\Lab\mu} &= \trival{k}{\previous \Lab \varphi}= \trival{k-1}{\Lab \varphi}\\ 
							&\stackrel{\text{IH}}{=} \trival{k-1}{\varphi} = \trival{k}{\previous \varphi}
	\end{align*}		
	\end{itemize}
	
	\item If $\mu = \DDia{\psi?} \varphi$. Note that, by definition, $\varphi,\psi \in FL(\mu)$. Moreover, it follows that

	\begin{align*}
	\trival{k}{\Lab\mu} &\stackrel{\eta(\mu)}{=} \trival{k}{\Lab{\psi} \wedge  \Lab\varphi}= \min\lbrace\trival{k}{\Lab{\psi}},\trival{k}{\Lab\varphi}\rbrace\\
						& \stackrel{IH}{\text{=}} \min\lbrace\trival{k}{\psi},\trival{k}{\varphi}\rbrace = \trival{k}{\psi \wedge \varphi}\\
						& \stackrel{\eqref{eq:valid:9}}{=} \trival{k}{\DDia{\psi?}\varphi}.
	\end{align*}
	
	\item If $\mu = \DBox{\psi?} \varphi$. Note that, by definition, $\varphi,\psi \in FL(\mu)$. Moreover, it follows that
	
	\begin{align*}
	\trival{k}{\Lab\mu} &\stackrel{\eta(\mu)}{=} \trival{k}{\Lab{\psi} \rightarrow  \Lab\varphi}\\
						 &= \begin{cases}
							2 & \text{ if } \trival{k}{\Lab{\psi}} \le \trival{k}{\Lab{\varphi}}\\
							\trival{k}{\Lab{\varphi}} & \text{ otherwise} 
							\end{cases} \\
						& \stackrel{\text{IH}}{=} \begin{cases}
											2 & \text{ if } \trival{k}{\psi} \le \trival{k}{\varphi}\\
											\trival{k}{\varphi} & \text{ otherwise } 
											\end{cases} \\
						&= \trival{k}{\psi \rightarrow \varphi} \stackrel{\eqref{eq:valid:10}}{=} \trival{k}{\DBox{\psi?} \varphi} \\
	\end{align*}
	
	\item If $\mu = \DDia{\rho_1 + \rho_2} \varphi$, let us take $\alpha = \DDia{\rho_1} \varphi$  and $\beta = \DDia{\rho_2} \varphi$
	\begin{align*}
	\trival{k}{\Lab\mu} & \stackrel{\eta(\mu)}{=} \trival{k}{\Lab \alpha \vee\Lab \beta}\\
						& = \max\lbrace\trival{k}{\Lab \alpha },\trival{k}{\Lab \beta}\rbrace.
	\end{align*}

\noindent We can apply the induction hypothesis on $\alpha$ and $\beta$ since $\alpha,\beta \in FL(\mu)$. Therefore we obtain 

	\begin{align*}
	\max\lbrace\trival{k}{\Lab \alpha },\trival{k}{\Lab \beta}\rbrace & \stackrel{\text{IH}}{=} & \max\lbrace\trival{k}{\alpha },\trival{k}{\beta}\rbrace\\
																	  & = & \trival{k}{\lbrace\alpha \vee \beta} \\
																	  & \stackrel{\eqref{eq:valid:2}}{=} & \trival{k}{\mu}.
	\end{align*}	

	\item If $\mu = \DBox{\rho_1 + \rho_2} \varphi$, let us take $\alpha = \DBox{\rho_1} \varphi$  and $\beta = \DBox{\rho_2} \varphi$.
	
	\begin{align*}
	\trival{k}{\Lab\mu} & \stackrel{\eta(\mu)}{=} \trival{k}{\Lab \alpha \wedge \Lab \beta}\\
	& = \min\lbrace\trival{k}{\Lab \alpha },\trival{k}{\Lab \beta}\rbrace.
	\end{align*}
	
	\noindent We can apply the induction hypothesis on $\alpha$ and $\beta$ since $\alpha,\beta \in FL(\mu)$. Therefore we obtain 

	\begin{align*}
	\min\lbrace\trival{k}{\Lab \alpha},\trival{k}{\Lab \beta}\rbrace & \stackrel{\text{IH}}{=} & \min\lbrace\trival{k}{\alpha },\trival{k}{\beta}\rbrace\\
	& = & \trival{k}{\lbrace\alpha \wedge \beta} \\
	& \stackrel{\eqref{eq:valid:1}}{=} & \trival{k}{\mu}.
	\end{align*}

	\item If $\mu = \DDia{\rho_1;\rho_2} \varphi$, let us consider $\alpha = \DDia{\rho_1}\DDia{\rho_2}\varphi$. Note that $\alpha \in FL(\mu)$. For this proof we will assume that $\eta(\mu) = \lbrace \Box(\Lab \mu \leftrightarrow \Lab\alpha)\rbrace$, which is equivalent to the assumption $\eta(\mu) = \eta(\alpha)$ (that is, the translation to $\mu$ can be directly replaced by the one of $\alpha$).
	
	\begin{align*}
	\trival{k}{\Lab\mu} & \stackrel{\eta(\mu)}{=} \trival{k}{\Lab \alpha}\stackrel{\text{IH}}{=} \trival{k}{\DDia{\rho_1}\DDia{\rho_2}\varphi}\\						
						& \stackrel{\eqref{eq:valid:4}}{=} \trival{k}{\DDia{\rho_1;\rho_2}\varphi}\\		
	\end{align*}	
	\item If $\mu = \DBox{\rho_1;\rho_2} \varphi$, let us consider $\alpha = \DBox{\rho_1}\DBox{\rho_2}\varphi$. Note that $\alpha \in FL(\mu)$. For this proof we will assume that $\eta(\mu) = \lbrace \Box(\Lab \mu \leftrightarrow \Lab\alpha)\rbrace$, which is equivalent to the assumption $\eta(\mu) = \eta(\alpha)$ (that is, the translation to $\mu$ can be directly replaced by the one of $\alpha$).
	\begin{align*}
		\trival{k}{\Lab\mu} & \stackrel{\eta(\mu)}{=} \trival{k}{\Lab \alpha}\stackrel{\text{IH}}{=} \trival{k}{\DBox{\rho_1}\DBox{\rho_2}\varphi}\\						
							& \stackrel{\eqref{eq:valid:3}}{=} \trival{k}{\DBox{\rho_1;\rho_2}\varphi}\\		
\end{align*}	

	\item For $\mu = \DDia{\rho^*} \varphi$, we distinguish two different cases depending on $k$
		\begin{itemize}
				\item if $k = \lambda -1$ we use the second formula in $\eta(\mu)$ to conclude
				
				\begin{align*}
				\trival{\lambda-1}{\Lab\mu}  & \stackrel{\eta(\mu)}{=} \trival{\lambda-1}{\Lab\varphi}  \stackrel{\text{IH}}{=}\trival{\lambda-1}{\varphi}\\
				& \stackrel{\eqref{eq:valid:8}}{=} \trival{\lambda-1}{\DDia{\rho^*}\varphi}.
				\end{align*}
				
				\item if $0 \le k < \lambda -1$, let us take $\alpha =\DDia{\rho} \DDia{\rho^*}\varphi $. By definition $\alpha, \varphi \in FL(\mu)$. Therefore, $\trival{k}{\Lab \varphi} = \trival{k}{\varphi}$ and $\trival{k}{\Lab\alpha} = \trival{k}{\alpha}$. With this we can use the first formula in $\eta(\mu)$ as follows:
				
				\begin{align*}
				        \trival{k}{\Lab\mu} &\stackrel{\eta(\mu)}{=} \trival{k}{\Lab\varphi \vee \Lab \alpha} \\ 
				        &=\max\lbrace\trival{k}{\Lab\varphi},\trival{k}{\Lab \alpha}\rbrace\\
					    &\stackrel{\text{IH}}{=}\max\lbrace\trival{k}{\varphi},\trival{k}{\alpha}\rbrace\\
					    &=\trival{k}{\varphi \vee \DDia{\rho}\DDia{\rho^*}\varphi}\\
					    &\stackrel{\eqref{eq:valid:7}}{=} \trival{k}{\mu}\\
				\end{align*}				
		\end{itemize}

	\item For $\mu = \DBox{\rho^*} \varphi$, we distinguish two different cases depending on $k$
	\begin{itemize}
		\item if $k = \lambda -1$ we use the second formula in $\eta(\mu)$ to conclude
		
		\begin{align*}
		\trival{\lambda-1}{\Lab\mu}  & \stackrel{\eta(\mu)}{=} \trival{\lambda-1}{\Lab\varphi}  \stackrel{\text{IH}}{=}\trival{\lambda-1}{\varphi}\\
		& \stackrel{\eqref{eq:valid:6}}{=} \trival{\lambda-1}{\DDia{\rho^*}\varphi}.
		\end{align*}
		
		\item if $0 \le k < \lambda -1$, let us take $\alpha =\DBox{\rho} \DBox{\rho^*}\varphi $. By definition $\alpha, \varphi \in FL(\mu)$. Therefore, $\trival{k}{\Lab \varphi} = \trival{k}{\varphi}$ and $\trival{k}{\Lab\alpha} = \trival{k}{\alpha}$. With this we can use the first formula in $\eta(\mu)$ as follows:
		
		\begin{align*}
		\trival{k}{\Lab\mu} &\stackrel{\eta(\mu)}{=} \trival{k}{\Lab\varphi \wedge \Lab \alpha} \\ 
		&=\min\lbrace\trival{k}{\Lab\varphi},\trival{k}{\Lab \alpha}\rbrace\\
		&\stackrel{\text{IH}}{=}\min\lbrace\trival{k}{\varphi},\trival{k}{\alpha}\rbrace\\
		&=\trival{k}{\varphi \wedge \DBox{\rho}\DBox{\rho^*}\varphi}\\
		&\stackrel{\eqref{eq:valid:5}}{=} \trival{k}{\mu}\\
		\end{align*}				
	\end{itemize}								
\end{enumerate}
\end{proofof}


\end{document}